\documentclass[final,5p,times,twocolumn,authoryear,nopreprintline]{elsarticle}

\usepackage{amssymb}

\usepackage[table, dvipsnames]{xcolor}
\usepackage{soul}
\usepackage{amsmath}
\usepackage{algorithm}
\usepackage[noend]{algpseudocode}
\usepackage{tabularx}
\usepackage{natbib}
\usepackage{caption}
\usepackage{subcaption}
\usepackage{float}
\usepackage{multirow}
\usepackage{dblfloatfix}
\usepackage{hyperref}
\usepackage{pifont}
\usepackage{pgfplots}
\pgfplotsset{compat=1.17}

\journal{Elsevier}

\begin{document}

\begin{frontmatter}



\title{Test-time adaptation for geospatial point cloud semantic segmentation with distinct domain shifts}


\author[polyu]{Puzuo Wang}
\ead{puzuo.wang@connect.polyu.hk}

\author[tum,geobim]{Wei Yao\corref{cor1}}
\ead{wei.yao@ieee.org}

\author[polyu]{Jie Shao}

\author[polyu]{Zhiyi He}

\address[polyu]{Dept. of Land Surveying and Geo-Informatics, The Hong Kong Polytechnic University, Hong Kong}
\address[tum]{School of Engineering and Design, Technical University of Munich, Munich, 80333, Germany}
\address[geobim]{GeoBIM \& GeoNexus Intelligence, Munich, 80997, Germany}


\cortext[cor1]{Corresponding author}


\begin{abstract}
Domain adaptation (DA) techniques aim to close the gap between source and target domains, enabling deep learning models to generalize across different data shift paradigms for point cloud semantic segmentation (PCSS). Among emerging DA schemes, test-time adaptation (TTA) facilitates direct adaptation of a pre-trained model to unlabeled data during the inference stage without access to source domain data and need for additional training process, which mitigates data privacy concerns and removes the requirement for substantial computational power. To fill the gap of leveraging TTA for geospatial PCSS, we introduce three typical domain shift paradigms in handling geospatial point clouds and construct three practical adaptation benchmarks, including photogrammetric point clouds to airborne LiDAR, airborne LiDAR to mobile LiDAR, and synthetic to mobile LiDAR. Then, a TTA method is proposed by exploiting the domain-specific knowledge embedded within the batch normalization (BN) layers. Given the pre-trained model, BN statistical information is progressively updated by fusing the statistics of each testing batch. Furthermore, we develop a self-supervised module to optimize the learnable BN affine parameters. Information maximization is used to generate confident and category-specific predictions, and reliability constrained pseudo-labeling is further incorporated to create supervisory signals. Extensive experimental analysis demonstrates that our proposed method significantly improves classification accuracy compared to directly applying the inference by up to 20\% in terms of mIoU, which not only outperforms other popular counterparts but also maintains a high efficiency while avoiding retraining. In an adaptation of photogrammetric  (SensatUrban) to airborne (Hessigheim 3D), our method achieves a mIoU of 59.46\% and an OA of 85. 97\%.

\end{abstract}




\end{frontmatter}


\section{Introduction}
\label{sec:intro}

Semantic segmentation of 3D point clouds plays a vital role in photogrammetry and remote sensing. By categorizing points with meaningful labels, semantic information supports various downstream applications that demand an in-depth understanding of 3D scenes, including land cover surveys \citep{HUANG202062}, 3D urban reconstructions \citep{10414143}, and forest inventories \citep{YAO2012368}.

\begin{figure*}[h]
\centering
\includegraphics[width=\linewidth]{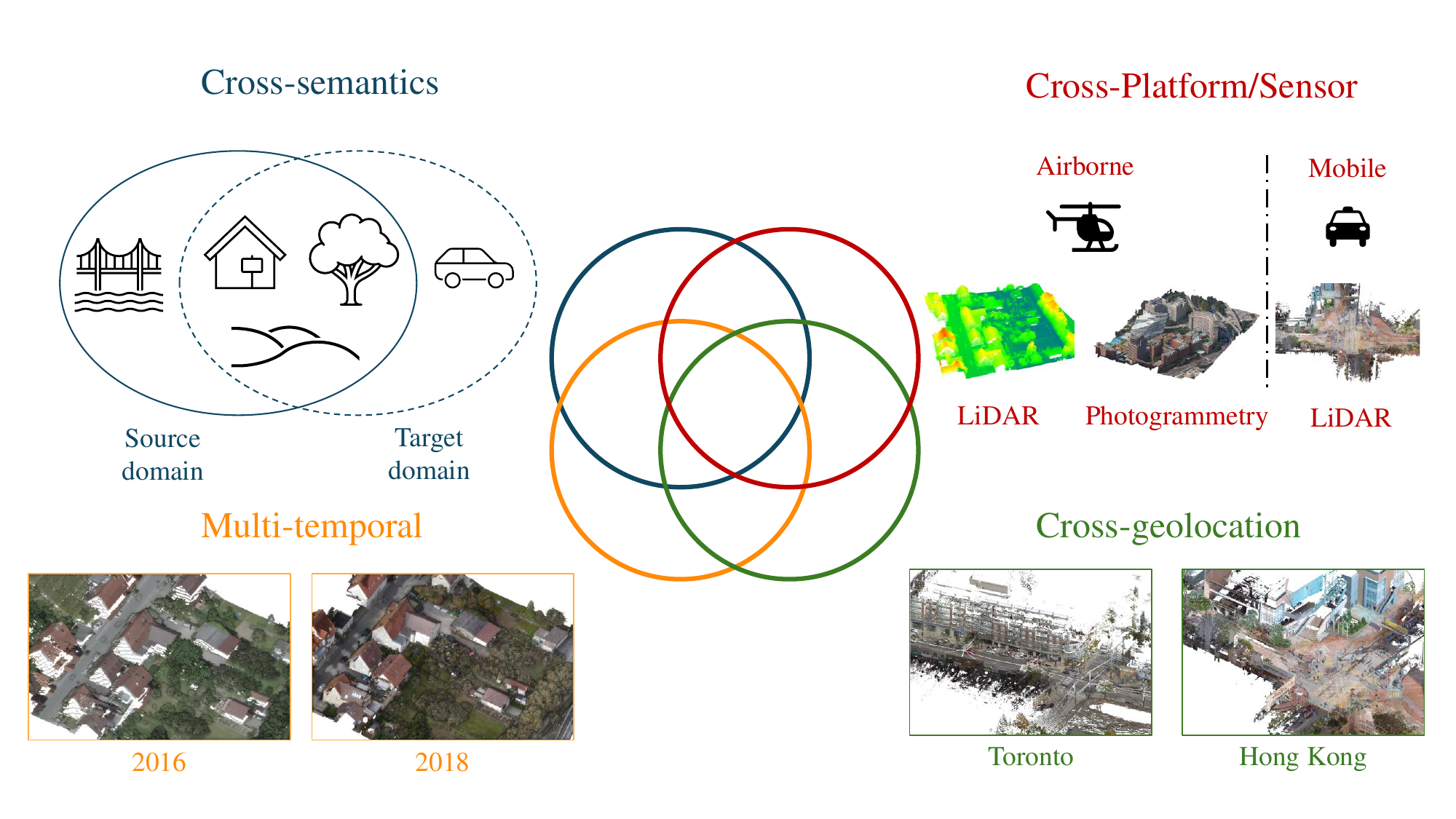}
\caption{Typical domain shifts in handling geospatial point clouds. }
\label{fig:ds}
\end{figure*}

In recent years, deep learning has emerged as a leading approach for this task, leveraging neural networks to extract high-level and distinct representations from raw point cloud data. Nevertheless, these approaches usually rely on numerous precise point-wise annotations to achieve favorable results, involving extensive and expensive labeling efforts. Point-wise annotations entail assigning a semantic label to each point in the point cloud, making it a difficult and time-intensive task. To address this issue, some researchers have turned to weakly supervised learning to reduce the annotation burden \citep{LIN202279,WANG2022237,WANG202389}. Weakly supervised learning seeks to train models using less detailed or noisier labels while still achieving performance comparable to fully supervised methods. Despite this, most of these models are designed and evaluated on specific 3D benchmarks, often neglecting generalization and transferability across various 3D domains. Different 3D domains may exhibit varying distributions, scales, densities, and noise levels in point clouds, leading to a marked performance decline when applying a model trained in one domain to another. Figure \ref{fig:ds} summarizes typical domain shifts in geospatial point clouds. Transfer learning aims to utilize the knowledge obtained by a trained model for related tasks, thereby addressing the issue of domain shifts. Given that the task variation in this study is due to differences in data structures, we use domain adaptation to describe this technological approach.

The research area of domain adaptation (DA) has become vital in machine learning, focusing on the problem of transferring knowledge from a source domain to a different target domain. In recent years, significant advancements have been achieved in addressing the growing complexities of DA \citep{10.5555/3045118.3045244,8099799,8954439}. Initial strategies utilized labeled data from both the source and target domains to learn domain-invariant features that convey transferable knowledge across domains. However, in practical computer vision scenarios, labeled target data is often limited or too costly to curate. As a result, unsupervised domain adaptation (UDA) has garnered considerable interest as a prominent DA branch. In UDA, models are trained on labeled source data and adapted to the target domain with only unlabeled target samples. Generative adversarial networks (GANs) have been extensively used in UDA approaches, aligning source and target distributions into a common feature space with minimized discrepancy \citep{8099799,pmlr-v80-hoffman18a}. A more challenging situation is found in source-free domain adaptation (SFDA), where only a pre-trained source model and target data are available for adaptation. This scenario is driven by cases where accessing source domain data is impractical due to privacy or confidentiality issues. Without source data access, some SFDA techniques have employed GANs to generate target-style training samples \citep{9157645} or isolate source-resembling target samples \citep{9710549} for guiding the target domain. Additionally, self-supervised methods have been broadly explored in SFDA, including entropy minimization \citep{10.5555/3524938.3525498}, contrastive learning \citep{ijcai2021p402}, and pseudo-labeling \citep{9578923}.

\begin{table}[t]
\caption{The comparison of different adaptation settings}
\label{tab:adapt}
\centering
\footnotesize
\setlength\tabcolsep{5pt}%
\begin{tabularx}{\linewidth}{Xccc}
\hline
Setting & Source data & Target data & Training \\
Domain adaptation & $x^s, y^s$ & $x^t, y^t$ & \ding{51} \\ 
Unsupervised domain adaptation & $x^s, y^s$ & $x^t$ & \ding{51} \\   
Source-free domain adaptation & - & $x^t$ & \ding{51} \\ 
Test-time adaptation & - & $x^t$ & \ding{55} \\ 
\hline
\end{tabularx}
\end{table}

In practical applications, an effectively trained model is expected to deliver precise classification results without needing further training, despite variations in data characteristics. This requirement has spurred the development of test-time adaptation (TTA), a concept centered around adjusting pre-trained models to manage novel test conditions and distributions during the inference stage. Due to its high efficiency, low computational demands, and capability to circumvent potential security threats linked to the source, TTA has garnered significant interest recently. Table~\ref{tab:adapt} shows the comparison of different adaptation settings. Faced with constrained resources, the researchers explored the viability of utilizing batch normalization (BN) \citep{10.5555/3045118.3045167} data for TTA tasks. Introduced in 2015, BN has established itself as a standard element in deep networks, computing the mean and variance of activations within a mini-batch and normalizing them. This process aids in mitigating internal covariate shift and accelerating convergence during training. Both experimental results and theoretical insights suggest that BN layer statistics capture data domain traits \citep{li2017revisiting,NEURIPS2020_85690f81}, making it an efficient mechanism to tackle domain shift problems. While one straightforward approach is to substitute the original BN statistics with those of the current test batch, some methods have merged BN data from both source and target domains \citep{9879821,lim2023ttn}. Moreover, self-supervision has supported adaptation by establishing auxiliary constraints and conducting test-time backpropagation. The widely used algorithms in this realm are akin to those employed in SFDA tasks.

In the domain of PCSS, TTA has received limited exploration \citep{10.1007/978-3-031-19827-4_33,9879729}. A possible explanation for the limited research in this area is the absence of a standardized TTA benchmark. When it comes to geospatial point clouds, the diversity in domain characteristics and label spaces—stemming from variations in regional attributes and acquisition methods—complicates the implementation of TTA. To facilitate TTA for geospatial point clouds and ensure fair comparisons, it is crucial to design meaningful and practical benchmarks. By evaluating existing public geospatial point cloud benchmarks, we have meticulously chosen several datasets for TTA following a comparison of label distribution similarities. We propose three key adaptation pathways, considering aspects such as area coverage and labeling complexity: photogrammetric to airborne laser scanning (ALS) data, ALS to mobile laser scanning (MLS) data, and synthetic to MLS data.

In pursuit of effective TTA, we propose to adapt the model through the optimization of BN layers, as these layers encapsulate domain-specific knowledge. Specifically, we introduce a progressive batch normalization module (PBN) to adapt the BN statistical information. Building upon a pre-trained model, the PBN module progressively updates the statistical information by incorporating the statistics calculated from each testing batch. Furthermore, we integrate a self-supervised method to optimize the BN affine parameters, including information maximization and pseudo-labeling. The proposed self-supervised learning exploits implicit semantic information in the target domain to enhance generalization performance, thus constructing a holistic model adaptation pipeline. This synergistic approach enables the effective transfer of domain-specific knowledge encapsulated within the BN layers, facilitating the adaptation of pre-trained models to unseen test conditions during inference time. Our main contributions are summarized as follows.

\begin{itemize}[leftmargin=*]

\item[$\bullet$] We highlight the importance of test-time adaptation for geospatial point cloud semantic segmentation, and construct three practical adaptation pathways/paradigms, encompassing  photogrammetric to ALS, ALS to MLS, and Synthetic to MLS adaptations. 

\item[$\bullet$] A method for updating BN statistical information is introduced to adapt the pre-trained model to the target domain, which is accomplished by progressively merging the BN statistics of each testing batch through an exponential moving average.

\item[$\bullet$] A self-supervised strategy is developed to optimize learnable BN parameters. Information maximization is employed to produce confident and category-specific predictions, and a reliability-constrained pseudo-labeling scheme considering entropy confidence and contrastive consistency is further incorporated.

\item[$\bullet$] Experimental findings based on designed benchmarks for three adaptation paradigms show that our approach yields a notable performance improvement over direct inference using the pre-trained model.

\end{itemize}

The remainder of the study is structured as follows. Section \ref{sec:re} provides a systematic review of UDA for PCSS and TTA task. Section \ref{sec:method} presents a detailed description of the proposed methodology. Section \ref{sec:exp} described the datasets used and the implementation details. In sections \ref{sec:res} and \ref{sec:dis}, we perform an extensive experimental study to evaluate and analyze the effectiveness of the proposed method. Finally, Section \ref{sec:con} concludes the study and discusses future research directions.

\section{Related works}
\label{sec:re}

\subsection{Unsupervised domain adaptation for point cloud semantic segmentation}

UDA seeks to enhance a model's performance in a target domain that lacks labeled data by utilizing information from a source domain with abundant label information. The primary challenge in UDA for PCSS is the acquisition of domain-invariant features that can effectively generalize from the source domain to the target domain. To tackle this issue, several methods, including adversarial learning, self-training, and feature alignment, have been proposed.

Using synthetic data is a prevalent strategy for this task, as large-scale and varied synthetic datasets can be generated with ease, and the process is generally more cost-effective and faster than manually annotating real-world data. Research has often concentrated on effective synthetic-to-real translation. An early work, SqueezeSegV2 \citep{10.1109/ICRA.2019.8793495}, investigated UDA for road-object segmentation by learning intensity information from synthetic point clouds through a pre-trained intensity rendering network. To enhance generalization performance, geodesic correlation alignment was employed to minimize the output distribution distance between the two domains, with BN parameters progressively adjusted layer by layer. ePointDA \citep{ePointDA} enhanced the SqueezeSegV2 network architecture by substituting all standard convolutions and the final conditional random field with aligned spatially adaptive convolutions and combining BN with instance normalization. Additionally, ePointDA addressed domain shifts at both the pixel and feature levels. Due to the absence of annotations in the target domain, auxiliary tasks were frequently utilized for cross-domain feature alignment. \citet{xiao2022transfer} assembled a synthetic dataset named SynLiDAR and used GAN to create the appearance translation module to convert synthetic point clouds to real ones and the sparsity translation module for projected depth images. In \citet{10.1109/ICRA46639.2022.9811654}, 2D range views were used to enhance feature learning by completing randomly omitted columns and matching the sparsity levels between domains. A gated adapter module (GA) was also incorporated to learn domain-specific details. In contrast to using projected 2D images, \citet{10204049} devised the Adaptive Spatial Masking module to replicate the noise in real LiDAR data and diminish domain shifts.

Alignment of cross-domain features is crucial for domain adaptation in real-world scenarios. xMUDA \citep{9157477} utilized the complementary information present in 2D images and 3D point clouds to tackle the domain shift issue. It presented a cross-modal learning framework that ensures consistency between the predictions of both modalities through mutual imitation. LiDARNet \citep{10.1109/ICRA48506.2021.9561255} introduced a domain adaptation model featuring a two-branch structure designed to extract both domain-specific features and domain-shared features. Domain-shared features were further used to predict boundary information as an auxiliary task to improve predictions. As outlined in \citet{9272684}, point-level and set-level losses were proposed to align local geometric details and the overall distribution between domains. PolarMix \citep{xiao2022polarmix} discovered that data augmentation could mitigate domain gaps by mixing LiDAR scans along the scan direction.

Given access to data from both the source and target domains, UDA methods primarily concentrate on aligning the two domains for adaptation. In contrast, our approach targets a more feasible and demanding adaptation strategy, aiming to enhance classification accuracy on the target domain by utilizing only a pre-trained model.

\subsection{Test-time adaptation}
TTA shows great promise in improving model performance and robustness in changing environments. Drawing from existing research, TTA methods can be classified into BN and self-supervised learning. Up to now, research has predominantly centered on 2D image classification.

\subsubsection{Batch normalization}

As a transformative technique that has significantly impacted deep learning training, batch normalization (BN) \citep{10.5555/3045118.3045167} tackles the internal covariate shift issue by standardizing layer inputs, stabilizing the training process and producing better outcomes. Given BN's superior impact on model robustness, \citet{NEURIPS2020_85690f81} leveraged the idea of covariate shift adaptation to fine-tune models for target distributions with corrupted data. By simply substituting the BN statistics computed on clean training data with those obtained from corrupted test data, improvements in robustness and convergence were seen across different out-of-distribution scenarios. DUA \citep{9879821} further utilized BN layer statistics to dynamically adapt a model to new data distributions. This was achieved by updating the mean and variance estimates of the BN layers online using a small portion of unlabeled data from the target domain. A similar approach was suggested for the UDA task \citep{li2017revisiting}, where the BN statistics were updated using a weighted average of statistics from the source and target domains. TTN \citep{lim2023ttn} also dynamically blends the source and test batch statistics for each BN layer depending on its sensitivity to changes in the domain. This sensitivity was assessed during a post-training phase, which calculated a gradient distance score between the clean input and its augmented version. In contrast to dynamically updating the BN statistics, \citet{10204752} proposed Dynamically Instance-Guided Adaptation (DIGA) for image semantic segmentation, where BN parameters were mixed from the source and each test sample.

\subsubsection{Self-supervised learning}

In addition to BN adaptation, self-supervised learning is frequently utilized in the TTA task to achieve performance gains. TENT \citep{wang2021tent} is the pioneering work that emphasizes the TTA task. It is built upon the insight that lower entropy signifies greater confidence and potentially more precise predictions. Therefore, TENT reduced the entropy of the model predictions as an indicator of this confidence. MEMO \citep{NEURIPS2022_fc28053a} additionally enforced regularization on the marginal entropy of model's predictions by utilizing different data augmentations on the test inputs. AdaContrast \citep{9880363} utilized self-supervised contrastive learning along with online pseudo-labeling to enhance feature learning for the target domain during test-time adaptation. Data augmentation and momentum models with gradual changes were used to form positive and negative pairs, while pseudo-labels were generated online and refined through soft voting among their nearest neighbors in the target feature space. \citet{10.1007/978-3-031-19827-4_26} presented a method for shift-agnostic weight regularization that manages how model parameters are updated individually for each layer by calculating the cosine similarity between gradient vectors from clean and augmented inputs. Furthermore, they proposed using contrastive learning with the nearest source prototypes to improve feature representations. Rather than modifying the model parameters, \citet{9879724} proposed a graph clustering approach that is optimized using the Laplacian Adjusted Maximum-likelihood Estimation (LAME) objective to enhance prediction accuracy. Given that machine perception systems function in dynamic and constantly evolving environments, various methods have explored solutions for continual adaptation. \citet{9880424} introduced Continual Test-Time Adaptation (CoTTA), which incorporates weight-averaged and augmentation-averaged predictions, as well as stochastic neuron restoration, to achieve long-term adaptation. \citet{pmlr-v162-niu22a} utilized an anti-forgetting regularizer to deter significant changes in model weights related to crucial features for in-distribution data.

To date, only a limited number of studies have explored TTA for PCSS, with the majority concentrating on autonomous scenarios. GIPSO \citep{10.1007/978-3-031-19827-4_33} utilized an auxiliary geometrically-informed encoder for propagating pseudo-labels to geometrically consistent distant regions, combined with a self-supervised temporal consistency constraint to enhance adaptation performance. In \citet{9879729}, both 2D and 3D modalities were used together for the fusion of cross-attention and self-attention features, while the generation of intramodal pseudo-labels and the refinement of intermodal pseudo-labels were introduced for joint learning. Our research aims to address the TTA issue between point cloud datasets across different modalities, which presents more significant domain shifts.

\section{Methodology}
\label{sec:method}

\begin{figure*}[!b]
\centering
\includegraphics[width=1\linewidth]{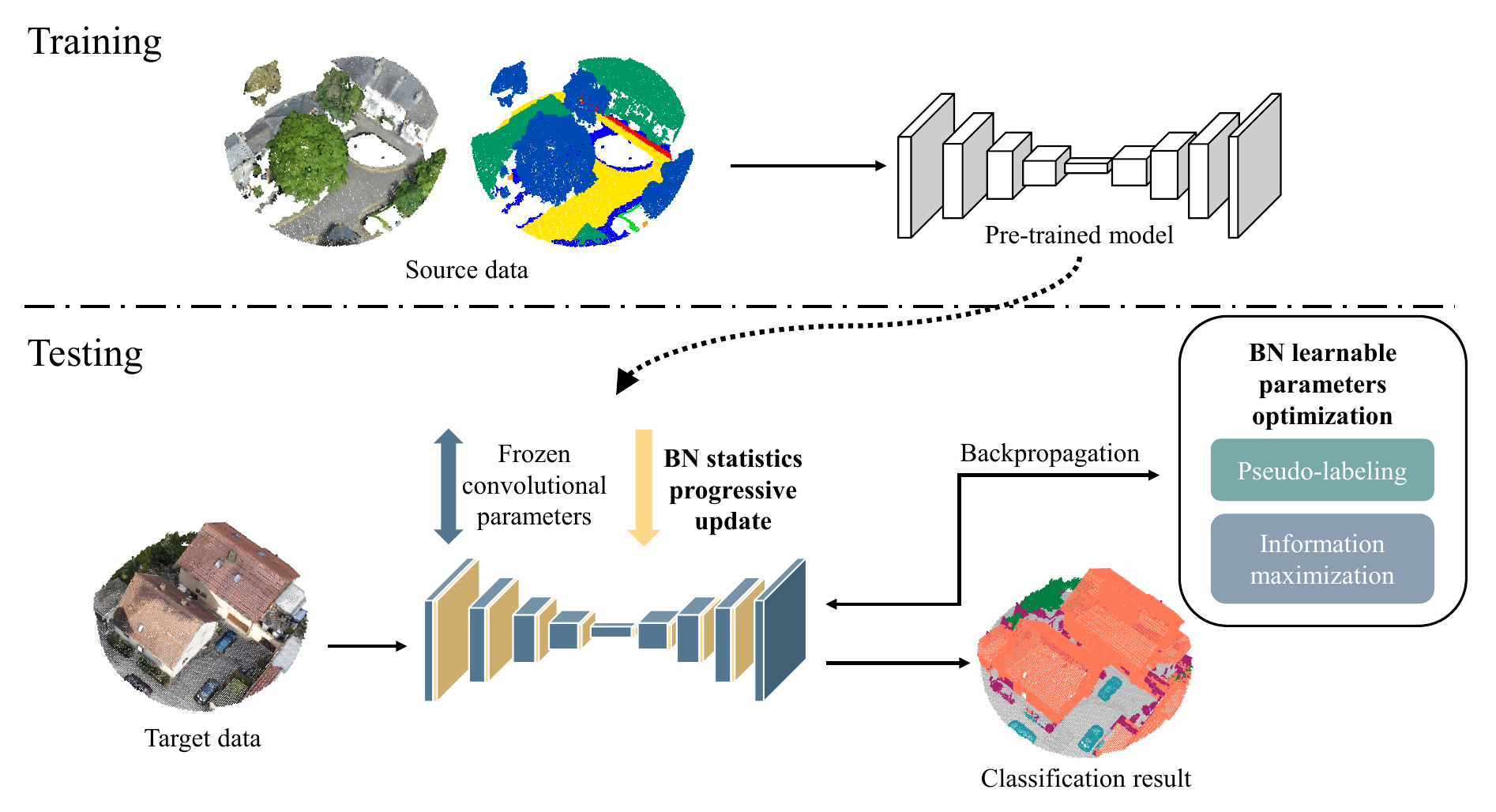}
\caption{Illustration of the proposed method for test-time adaptation. Following the training of a deep model on labeled source data, the pre-trained model is adapted to the target data through modifications to the BN layers during inference, which includes updating statistical information progressively and optimizing learnable parameters in a self-supervised manner.}
\label{fig:pipeline}
\end{figure*}

\subsection{Overview}

Given a pre-trained model $M$ in a source domain $D^s$, TTA aims to discover a method $f$ that adjusts $M$ to new data from the target domain $x^t \in D^t$, ensuring that the adjusted model $M^t = f(M, x^t)$ reduces the expected risk on the target distribution. Considering stability issues due to completely unlabeled target data, inspired by \citet{wang2021tent}, we directly use the BN layers of the pre-trained model, which possess domain-specific knowledge. Building on this, we introduce a progressive batch normalization method (PBN) to adjust BN statistical data and a self-supervised technique involving information maximization and pseudo-labeling to fine-tune BN affine parameters (see Fig.~\ref{fig:pipeline}). All other model parameters remain unchanged during testing.

\begin{figure*}[!b]
\centering
\begin{subfigure}{0.25\linewidth}
    \includegraphics[width=1.0\linewidth]{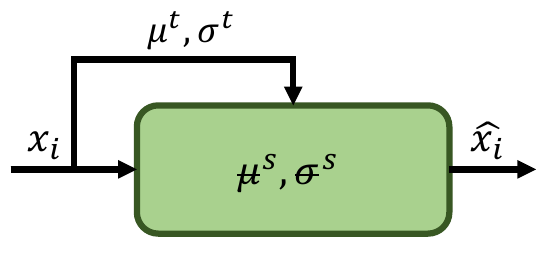}
    \caption{Adaptive BN}
\end{subfigure}
\hfill
\begin{subfigure}{0.25\linewidth}
    \includegraphics[width=1.0\linewidth]{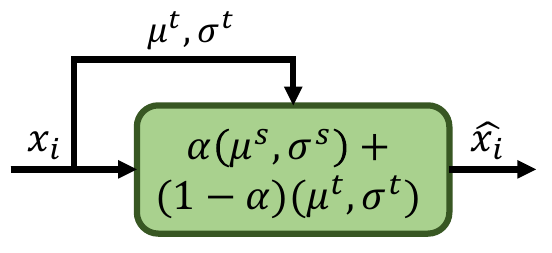}
    \caption{Weighted BN adaptation}
\end{subfigure}
\hfill
\begin{subfigure}{0.25\linewidth}
    \includegraphics[width=1.0\linewidth]{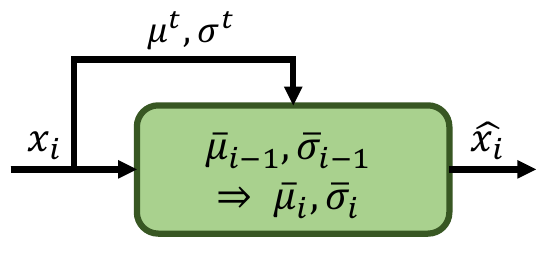}
    \caption{Progressive BN adaptation}
\end{subfigure}
\hfill
\caption{Comparison of different BN adaptation methods.}
\label{fig:bn}
\end{figure*}

\subsection{Progressive batch normalization}

Considering that BN information includes domain-specific knowledge, we propose a progressive batch normalization (PBN) technique to combine BN statistics from the source and target domains.

\subsubsection{Revisiting batch normalization in deep learning}

BN \citep{10.5555/3045118.3045167} is recognized as a vital technique in deep learning. It addresses the well-known issue of internal covariate shift, which considerably impedes the learning process and affects generalization. The internal covariate shift refers to the variation in the distribution of inputs to a layer within a deep neural network during training. Deep neural networks are trained in batches, and each batch possesses its own unique statistical properties, such as mean and variance. During training, as the weights and biases in previous layers are updated, the distribution of inputs to the subsequent layers can vary extensively. This causes multiple problems, including slower convergence, vanishing / explosive gradients, and decreased generalization.

To address this problem, BN standardizes the activations of each layer by removing the batch mean and scaling by the batch standard deviation. This process re-centers and rescales the inputs within various batches. Specifically, for a batch of inputs {$x_1$, $x_2$, ..., $x_m$} to a layer, BN calculates the mean $\mu$ and variance $\sigma^2$ across the batch:
\begin{equation}
\label{equ:mu&sigma}
\begin{aligned}
& \mu = \frac{1}{m}\sum_{i=1}^{m}x_i,
& \sigma^2 = \frac{1}{m}\sum_{i=1}^{m}(x_i - \mu)^2
\end{aligned}
\end{equation}
Then, the inputs are normalized by subtracting the mean and dividing by the square root of the variance, along with adding a small constant epsilon ($\epsilon$) for numerical stability:
\begin{equation}
\hat{x}_i = \frac{x_i - \mu}{\sqrt{\sigma^2 + \epsilon}}
\end{equation}
The normalized activations are then scaled and shifted using learnable parameters gamma ($\gamma$) and beta ($\beta$), respectively, ensure the network retain the representational capacity:
\begin{equation}
x'_i = \gamma\hat{x}_i + \beta
\end{equation}
The parameters $\gamma$ and $\beta$ are learned during the training process, allowing the network to adjust the scale and shift of the normalized activations as required.

\subsubsection{Progressive adaptation}
In DA tasks, the primary objective is to reduce prediction errors caused by cross-domain variations. Owing to the conceptual parallels between domain shifts and internal covariate shifts, scholars began utilizing BN data to address DA issues. While training on source datasets, batches are standardized considering the mean and variance of the present batch, and $\mu,\sigma$ are aggregated through a running average as the network processes multiple batches:
\begin{equation}
\begin{aligned}
& \bar{\mu}_i = (1-\rho)\cdot\bar{\mu}_{i-1} + \rho\cdot{\mu}_i \\
& \bar{\sigma}_i^2 = (1-\rho)\cdot\bar{\sigma}_{i-1}^2 + \rho\cdot{\sigma}_i^2
\end{aligned}
\end{equation}
where $i$ represents the training step number, and $\rho$ denotes the smoothing factor. During standard inference, $\bar{\mu}$ and $\bar{\sigma}$ are fixed to normalize the input because the averaged value over the entire training set is considered to encapsulate comprehensive domain statistical information. However, the mean and variance values $\mu^s$ and $\sigma^s$ calculated using the source data may not adequately reflect the domain characteristics of the target data when there are significant domain changes, often resulting in performance degradation. To address this, \citet{li2017revisiting} proposed a simple method called adaptive BN. Since BN ensures that a similar distribution is fed into each layer regardless of whether the data come from the source domain or the target domain, the mean and variance of the target batch, $\mu^t$ and $\sigma^t$, are directly used during inference, expressed as:
\begin{equation}
{x'_i}^t =  \gamma^s\frac{x_i^t - \mu^t}{\sqrt{(\sigma^t)^2 + \epsilon}} + \beta^s
\end{equation}
where $\mu^t,\sigma^t$ are calculated with Equ.~\ref{equ:mu&sigma}, and $\gamma^s,\beta^s$ are learned from source data. 

Considering the issue of unreliability due to the small test batch size, the methods examined using $\mu^s,\sigma^s$ as prior information and integrating them with $\mu^t,\sigma^t$ in BN layers. One type of approach involved computing a weighted sum of $\mu^s,\sigma^s$ and $\mu^t,\sigma^t$ for each test batch \citep{10204752}, while others updated $\mu^t,\sigma^t$ as the testing progressed \citep{9879821}. The comparison of various BN adaptations is shown in Fig.~\ref{fig:bn}. In this study, we contend that the latter approach is more suitable for our scenario. Large-scale PCSS often encompasses multiple highly homogeneous test batches. Therefore, although the weighted BN for each test batch can prevent accumulated errors in certain tasks, the homogeneity among test batches actually makes cumulative $\mu^t,\sigma^t$ beneficial for the adaptation process in our datasets. Another benefit of the cumulative value is its compatibility with self-supervised optimization techniques. Specifically, given a pre-trained model with $\mu^s,\sigma^s$ as $\bar{\mu}_0,\bar{\sigma}_0$, PBN incrementally updates $\bar{\mu},\bar{\sigma}$ during each test batch as:
\begin{equation}
\begin{aligned}
& \bar{\mu}_i = (1-\rho)\cdot\bar{\mu}_{i-1} + \rho\cdot\mu_i^t \\
& \bar{\sigma}_i^2 = (1-\rho)\cdot\bar{\sigma}_{i-1}^2 + \rho\cdot(\sigma_i^t)^2
\end{aligned}
\end{equation}
As indicated by \citep{9879821}, using a small $\rho$ effectively mitigates instability issues. While \citet{9879821} also proposed a variable $\rho$ to attain quicker convergence, our results show that a constant $\rho$ is enough to ensure good performance without the need for extra hyperparameter tuning.

\subsection{Learnable parameter adaptation with self-supervised learning}

Although adjusting $\mu$ and $\sigma$ helps address the domain shift issue and improves adaptation effectiveness, the parameters $\gamma$ and $\beta$ in the BN layers remain unchanged. To achieve complete BN adaptation, we refine $\gamma$ and $\beta$ using a self-supervised approach that incorporates information maximization (IM) and pseudo-labeling (PL).
\begin{figure*}[t!]
\centering
\includegraphics[width=0.95\linewidth]{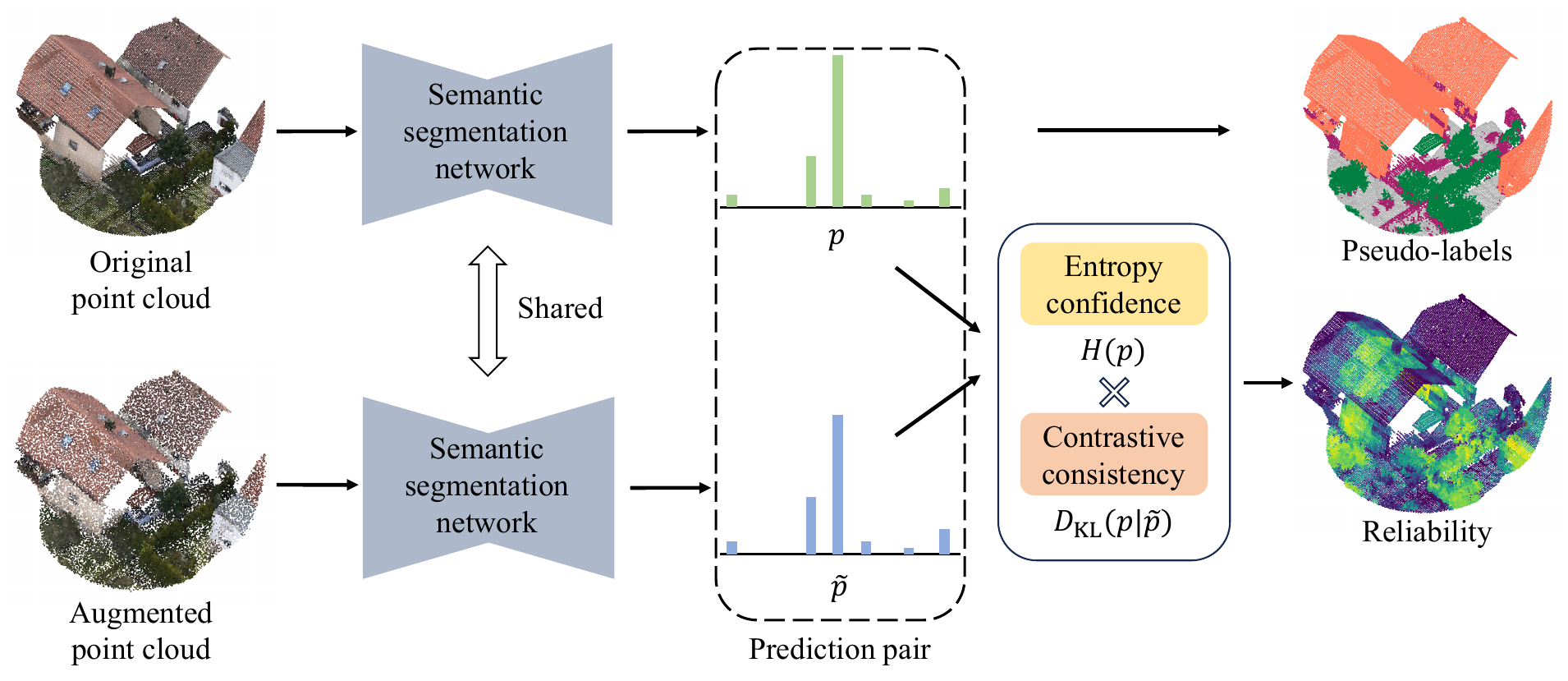}
\caption{The reliability constrained pseudo-labeling strategy, which jointly considers entropy-based confidence and contrastive consistency.}
\label{fig:pl}
\end{figure*}

\subsubsection{Information maximization}

A pioneering study introduced entropy regularization (ER) \citep{10.5555/2976040.2976107} to mitigate adaptation errors \citep{wang2021tent}. ER reduces high-entropy (uncertain) predictions on unlabeled data, which is a common approach in semi-supervised tasks. It pushes the model to make more confident (low-entropy) predictions and improve performance, as low entropy typically signifies high accuracy. ER refines the deep model by decreasing the Shannon entropy \citep{shannon} $H$ of the predicted probability, expressed as
\begin{equation}
\begin{aligned}
&\mathcal{L}_{\text{er}} = \frac{1}{N}\sum_{i}^N H_i, \\ &H_i=-\sum_{c}^{K}p_{ic}\log p_{ic}
\end{aligned}
\label{equ:er}
\end{equation}
where $p_{ic}$ represents the predicted probability of point $p_i$ belonging to the $c$-th category. Nevertheless, in the TTA scenario, the absence of annotation data prevents dependable supervision, making it prone to trivial solutions. For instance, $\mathcal{L}_{\text{er}}$ will drop to 0 if all points are assigned to a single class with 100\% probability. To preserve prediction diversity, inspired by \citet{10.5555/3524938.3525498}, we incorporate IM to fine-tune $\gamma$ and $\beta$. Specifically, in addition to $\mathcal{L}_{\text{er}}$, an additional diversity loss is added:

\begin{equation}
\mathcal{L}_{\text{div}} = -\sum_{c}^{K} \bar{p}_{c}\log \bar{p}_{c}
\end{equation}
where $\bar{p}_{c}=\frac{1}{N}\sum_{i}^N p_{ic}$ denotes mean predicted probability of $c$-th category. IM simultaneously minimizes $\mathcal{L}_{\text{er}}$ and $\mathcal{L}_{\text{div}}$ to optimize model parameters through $\mathcal{L}_{\text{im}} = \mathcal{L}_{\text{er}} + \mathcal{L}_{\text{div}}$.

\subsubsection{Pseudo-labeling with reliability constraint}
IM promotes the generation of confident and diverse category predictions. Yet, without label data, the model might still produce unstable outcomes, such as sensitivity to the learning rate. To mitigate these negative effects, we introduce pseudo-labeling \citep{lee2013pseudo} to provide supervision signals. Pseudo-labels $l^{\text{pl}}$ are the model's predictions on unlabeled data, acting as proxy true labels. Originally developed for semi-supervised learning, pseudo-labeling aids TTA tasks where annotations are lacking. It is crucial to consider the confirmation bias of pseudo labels. Therefore, we limit the impact of pseudo labels using contrastive consistency and entropy uncertainty (see Fig.~\ref{fig:pl}).

In DA problems, divergence in the data distribution between the source and target domains presents challenges in leveraging the target data for adaptation. Domain shifts result in noisy information and error propagation when self-supervised techniques are employed. An optimal adaptation process aims to minimize the impact of out-of-distribution data and fully utilize domain-general knowledge, propagating it to the entire target domain. Contrastive consistency is a widely used mechanism in DA tasks, as high consistency in contrastive predictions often signifies high reliability and domain-general information. Various methods, such as data augmentation \citep{9710018}, Bayesian models \citep{10.5555/3367471.3367576}, and dropout \citep{9008834}, are commonly used to build contrastive pairs. In this study, we enhance the initial point clouds and construct contrastive data pairs to assess prediction reliability. Specifically, we add random noise to the raw data $x$ to slightly perturb the original geometric structure, which has proven to be effective based on experimental results. This approach is motivated by the hypothesis that consistent predictions from contrastive data pairs represent domain-invariant information, which is beneficial for robust adaptation. $x$ and its augmented version $\tilde{x}$ are fed into the network to produce contrastive predicted probabilities ($p,\tilde{p}$). To harness the full potential of all pseudo-labels during model optimization, rather than using only prediction-consistent data, we estimate the significance of pseudo-labels through Kullback–Leibler (KL) divergence \citep{1320776d-9e76-337e-a755-73010b6e4b64}. KL divergence measures the similarity between two probability distributions ($p,\tilde{p}$), and is given by
\begin{equation}
D_\text{KL}(p||\tilde{p}) = \sum_{c}^K p_c\log  \left(\frac{p_c}{\tilde{p}_c}\right)
\label{equ:kl}
\end{equation}
By measuring the disparity between $p$ and $\tilde{p}$, $D_\text{KL}$ is used to estimate the reliability of pseudo-labels  throughout the optimization process. $D_\text{KL}$ is transferred to weight value $w^{\text{con}}=\exp(-D_\text{KL})$.

Meanwhile, predictions with high confidence and low uncertainty are more likely to be correctly classified, which can be seen as an effective indicator of reliability. We take the entropy $H$ to calculate the prediction confidence as 
\begin{equation}\label{equ:w_ent}
	 w^{\text{ent}}=1-\frac{H}{\log K}
\end{equation}
Next, we define the final reliability weight $w$ as the product of $w^{\text{con}}$ and $w^{\text{ent}}$, denoted as $w=w^{\text{con}}\cdot w^{\text{ent}}$. The loss function for reliability-constrained pseudo-labeling is then expressed as:\begin{equation}\label{equ:pl_r}
	\mathcal{L}_{\text{pl}} = - \frac{1}{\sum_{i}^N w_i} \sum_{i}^N w_i \sum_{c}^{K} y^{\text{pl}}_{ic}\log p_{ic}
\end{equation}
$y^{\text{pl}}_{ic}$ = 1 if $c$ equals to label $l^{\text{pl}}_i$, otherwise 0.

\begin{figure*}[b!]
\centering
\begin{subfigure}{1.0\linewidth}
    \includegraphics[width=1\linewidth]{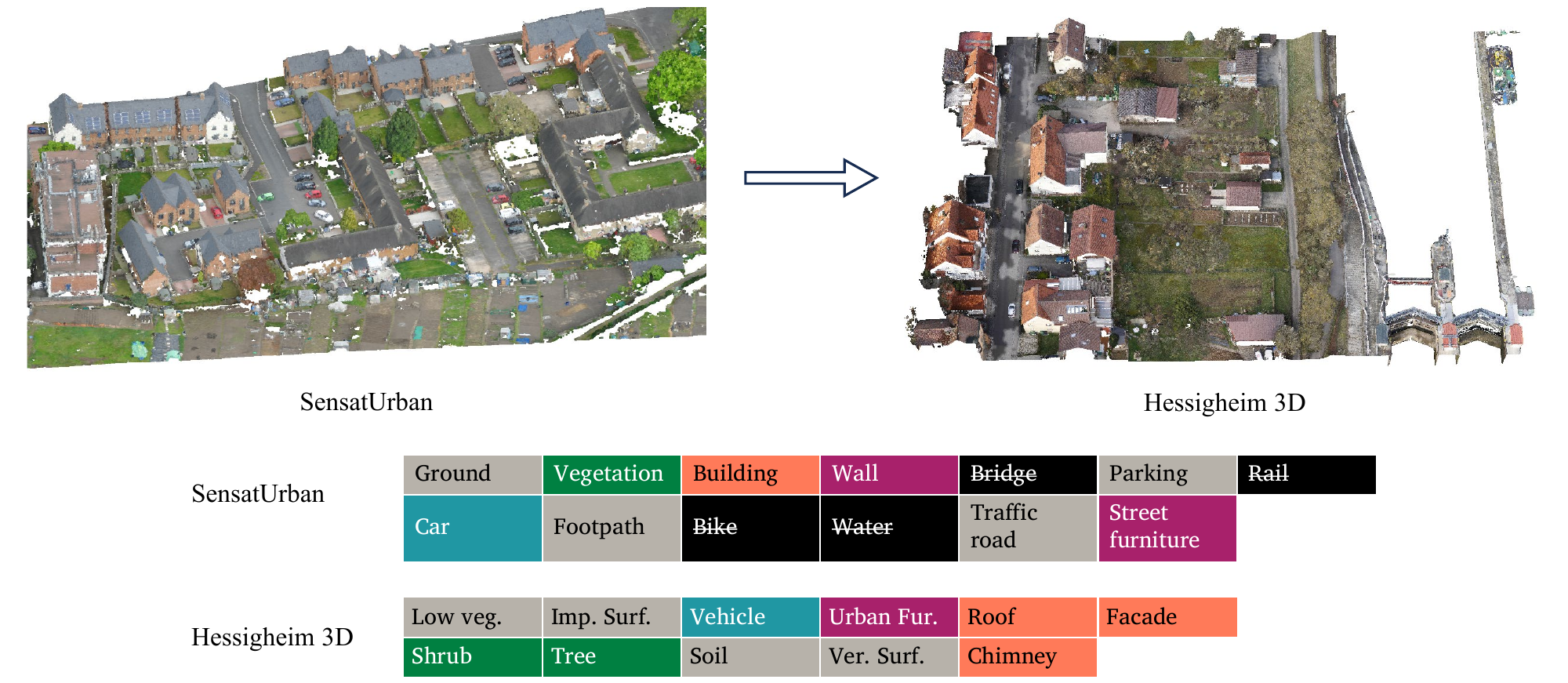}
    \caption{Photogrammetric to ALS data adaptation}
\end{subfigure}
\hfill
\begin{subfigure}{1.0\linewidth}
    \includegraphics[width=1\linewidth]{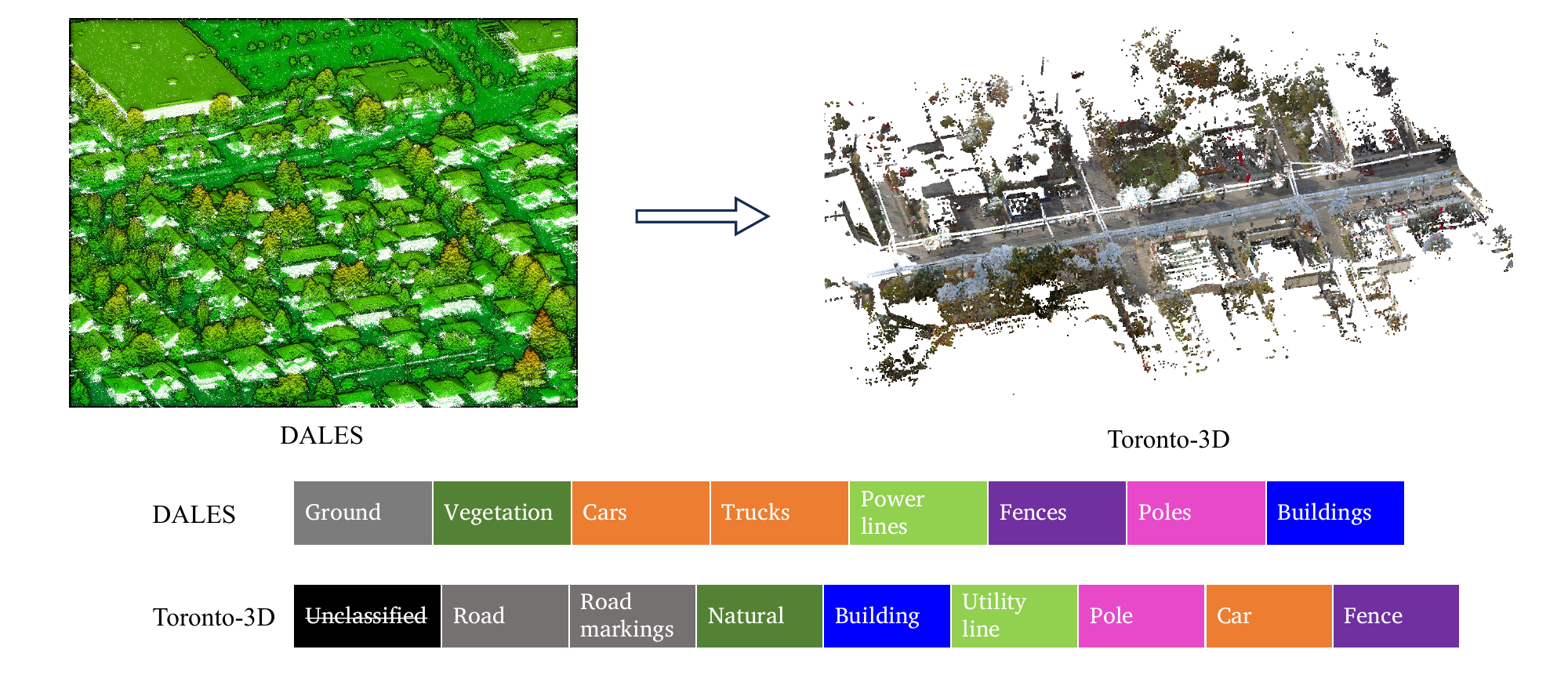}
    \caption{ALS to MLS data adaptation}
\end{subfigure}
\end{figure*}
\begin{figure*}[t]\ContinuedFloat
\centering
\begin{subfigure}{1.0\linewidth}
    \includegraphics[width=1\linewidth]{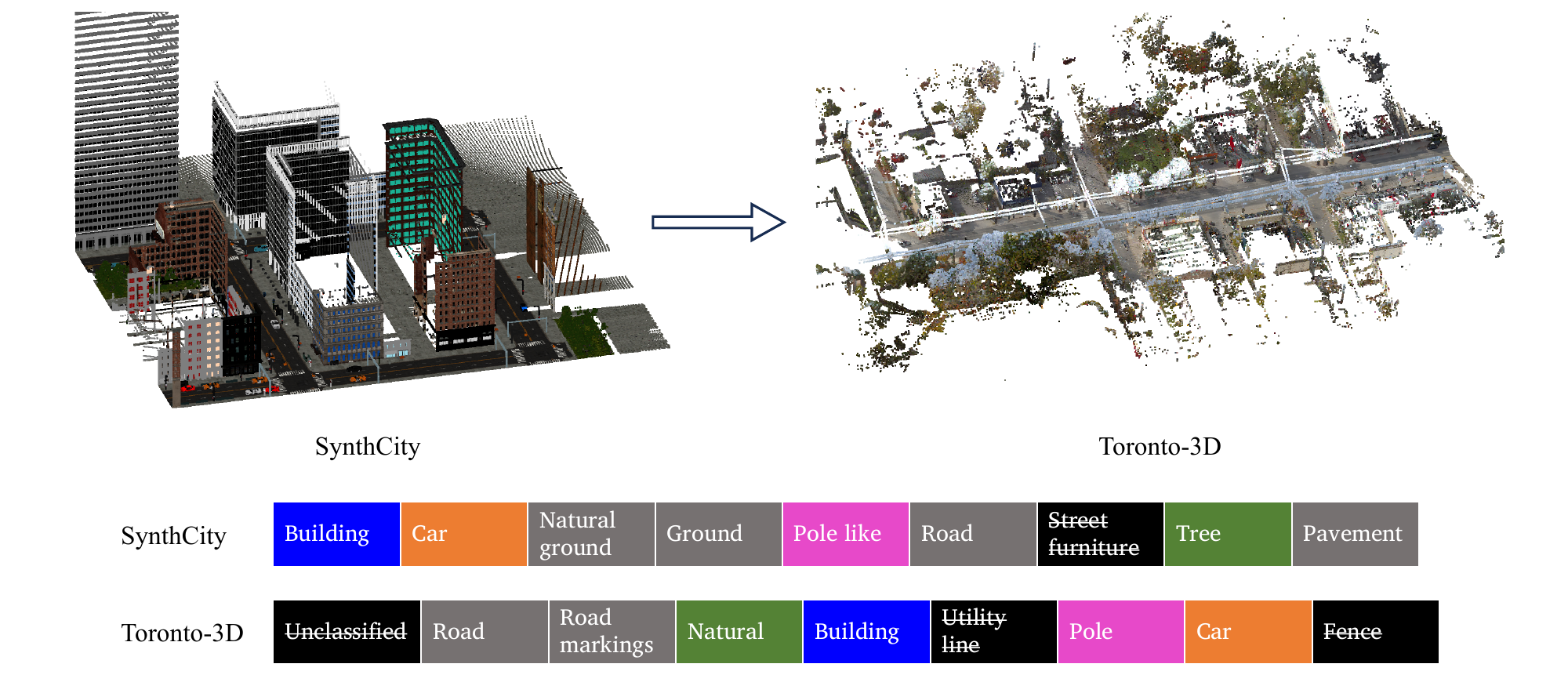}
    \caption{Synthetic to MLS data adaptation}
\end{subfigure}
\caption{Three real-world TTA benchmarks. Based on the category distribution among the datasets, evaluation merges classes sharing the same color, while black-colored categories are excluded from evaluation.}\label{fig:benchmark}
\end{figure*}

\section{Experiment}
\label{sec:exp}

\subsection{Dataset and benchmark}
Domain shifts in geospatial point clouds typically arise from various factors like location, acquisition platform, modality, and time. Considering the scale of the data, along with the costs of acquisition and labeling, as well as the category overlap across datasets, three practical adaptation scenarios are examined using 5 public point cloud datasets. These scenarios include adapting photogrammetric (SensatUrban \citep{10.1007/s11263-021-01554-9}) to ALS (Hessigheim 3D \citep{KOLLE2021100001}), ALS (DALES \citep{9150622}) to MLS (Toronto-3D \citep{9150609}), and synthetic (SynthCity \citep{griffiths2019synthcity}) to MLS (Toronto-3D). We start by describing the characteristics of these datasets.

\subsubsection{Dataset description}

\paragraph{SensatUrban}
SensatUrban is a photogrammetric point cloud dataset captured using unmanned aerial vehicles (UAVs). It contains nearly 3 billion points, covering an area of 7.6 $km^2$ across three cities in UK. Each point is meticulously labeled with one of 13 semantic categories.

\paragraph{DALES}
Dayton Annotated Laser Earth Scan (DALES) is a large-scale and diverse dataset that covers a massive collection of 40 densely labeled aerial scenes spanning various environments such as urban, suburban, rural, and commercial. The entire data spans 330 $km^2$ over the city of Surrey, Canada. Eight classes are manually annotated for study.

\paragraph{Hessigheim 3D (H3D)}
The dataset consists of a high-resolution LiDAR point cloud, featuring a density of around 800 points/$m^2$, complemented by an RGB image with a Ground Sampling Distance (GSD) between 2 and 3 centimeters. The area of study was situated in Hessigheim, Germany. For the experiments, data gathered in March 2018 was utilized, and the evaluation was carried out using the validation data. The analysis involved 11 predefined semantic categories.

\paragraph{Toronto-3D (T3D)}
  Toronto-3D (T3D) is a point cloud dataset obtained through MLS, tailored for studies related to urban highways. It encompasses about 1 $km$ of urban roadways in Toronto, Canada. This dataset contains approximately 78.3 million points and is classified into eight semantic categories. The file named ``L002" is used for evaluation purposes.

\paragraph{SynthCity}
SynthCity is a fully synthetic MLS dataset created programmatically with computer graphics software. It consists of an immense 367.9 million colorized points categorized into 9 categories.

\subsubsection{Benchmark}
In our practical TTA approach for PCSS, given the need for adaptation to varying category distributions, we develop and adjust the classification benchmark to facilitate viable adaptation experiments. The constructed TTA classification benchmark is depicted in Fig.~\ref{fig:benchmark}. When comparing source and target datasets, analogous classes are combined for evaluation, while dissimilar classes are excluded. Models are first trained on labeled source data, then assessed using a test set from the target domain.

\subsection{Implementation details}

Some configuration specifications are provided here. Given the high density of the raw data, we subsampled a subset to improve computational efficiency while maintaining the detailed structure. Due to variations in point density between datasets, we standardized the grid size to 0.2 meters for all datasets. We used overlapping spherical sub-clouds, uniformly set at 15 meters, for mini-batch creation. Each batch, during testing, is capped at 40,000 points. Once training is complete in the source domain, we leverage the pre-trained model for label classification in the target test dataset. To ensure a fair comparison, we maintain the standard inference efficiency, testing each point approximately three times. KPConv \citep{kpconv} is used as the backbone network, and we use the Adam optimizer with a learning rate of $10^{-4}$. All models are implemented in the PyTorch framework.

\begin{figure*}[b!]
\centering
\begin{subfigure}[b]{1.0\linewidth}
    \centering
    \includegraphics[width=0.45\linewidth]{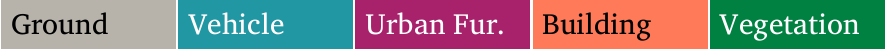}
\end{subfigure}
\hfill
\begin{subfigure}{0.32\linewidth}
    \centering
    \includegraphics[width=1\linewidth]{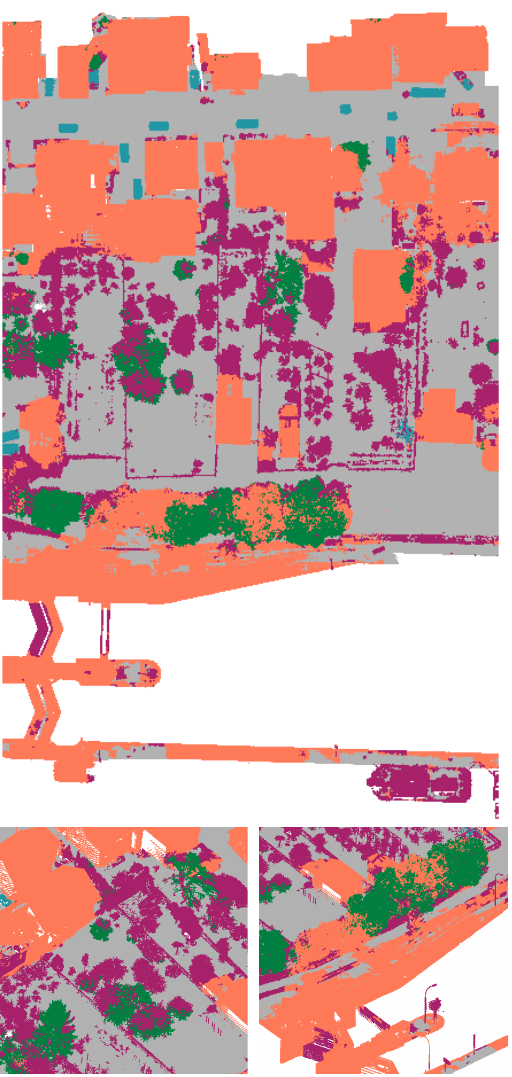}
    \caption{Source only}
\end{subfigure}
\hfill
\begin{subfigure}{0.32\linewidth}
    \centering
    \includegraphics[width=1\linewidth]{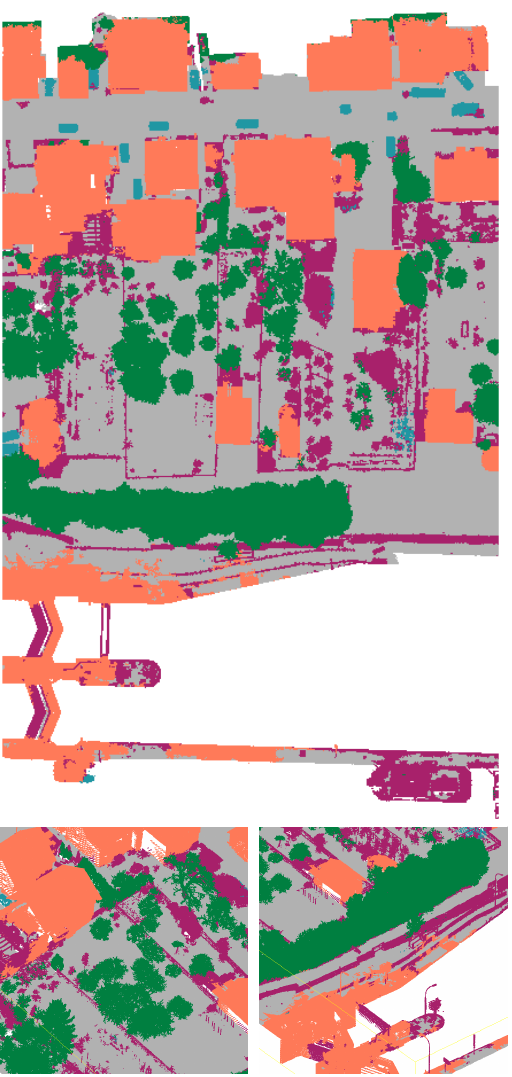}
    \caption{Our method}
\end{subfigure}
\hfill
\begin{subfigure}{0.32\linewidth}
    \centering
    \includegraphics[width=1\linewidth]{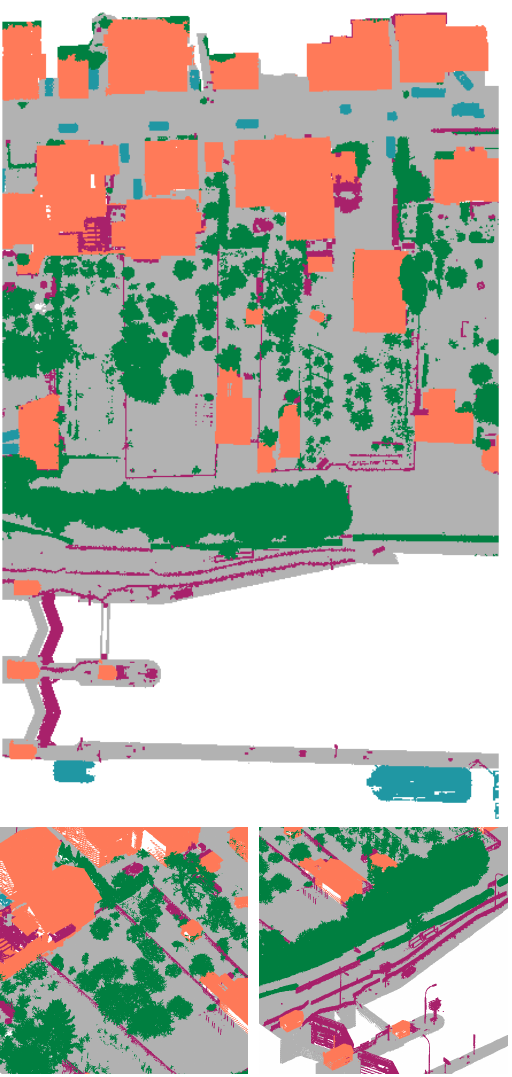}
    \caption{Ground truth}
\end{subfigure}
\caption{Classification result of H3D dataset with the model adapted from SensatUrban dataset.}
\label{fig:res_h3d}
\end{figure*}

\begin{table*}[b!]
\caption{Comparison of TTA methods on SensatUrban to H3D dataset (\%)}
\label{tab:h3d}
\centering
\small
\begin{tabularx}{\textwidth}{XXXXXXXX}
\hline
\multirow{2}{*}{Method} & \multicolumn{5}{c}{IoU} & \multirow{2}{*}{mIoU} & \multirow{2}{*}{OA} \\ 
\cline{2-6}  & Ground &  Vehicle & Urban Fur. & Building & Vegetation  \\ \hline                        
Source & 80.36 & 34.25 & 8.80 & 59.46 & 26.39 & 41.85 & 73.48 \\
AdaBN & 73.41 & 31.06 & 17.68 & 57.98 & 73.31 & 50.69 & 78.56 \\ 
TENT & 73.02 & 30.88 & 17.78 & 58.00 & 74.06 & 50.75 & 78.57  \\ 
DIGA & 85.82 & 11.38 & 10.82 & 67.03 & 52.08 & 45.43 & 78.11 \\
Ours  & 85.45 & 37.53 & 16.71 & 77.82 & 79.81 & 59.46 & 85.97 \\ 
\hline
\end{tabularx}
\end{table*}

\subsection{Evaluation metrics}
The overall accuracy (OA) and Intersection over Union (IoU) are used to evaluate the performance of our method. OA stands for the percentage of correctly predicted points, while IoU quantifies the overlap between predicted classification results and ground truth, expressed as:
\begin{equation}
IoU = \frac{{tp}}{{tp + fp + fn}}
\end{equation}
where $tp$, $fp$, and $fn$ are the true positives, false positives, and false negatives, respectively.

\section{Result}
\label{sec:res}

In this analysis, we examine the experimental results and compare our approach with several well-known adaptation methodologies, such as Source, AdaBN \citep{li2017revisiting}, TENT \citep{wang2021tent}, and DIGA \citep{10204752}. The Source serves as the baseline, which uses a traditional inference method without incorporating adaptation techniques. AdaBN substituted the source domain BN statistics with the batch-wise target data statistics, while TENT additionally used ER to achieve predictions with high certainty. DIGA is an innovative back-free TTA technique aimed at semantic segmentation, developed through weighted BN adaptation and semantic adaptation employing class prototypes.

\begin{figure*}[b!]
\centering
\begin{subfigure}[b]{1.0\linewidth}
    \centering
    \includegraphics[width=0.72\linewidth]{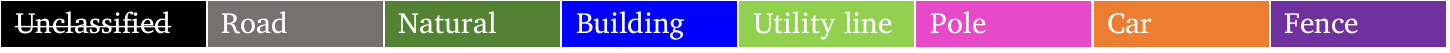}
\end{subfigure}
\hfill
\begin{subfigure}{0.32\linewidth}
    \centering
    \includegraphics[width=1\linewidth]{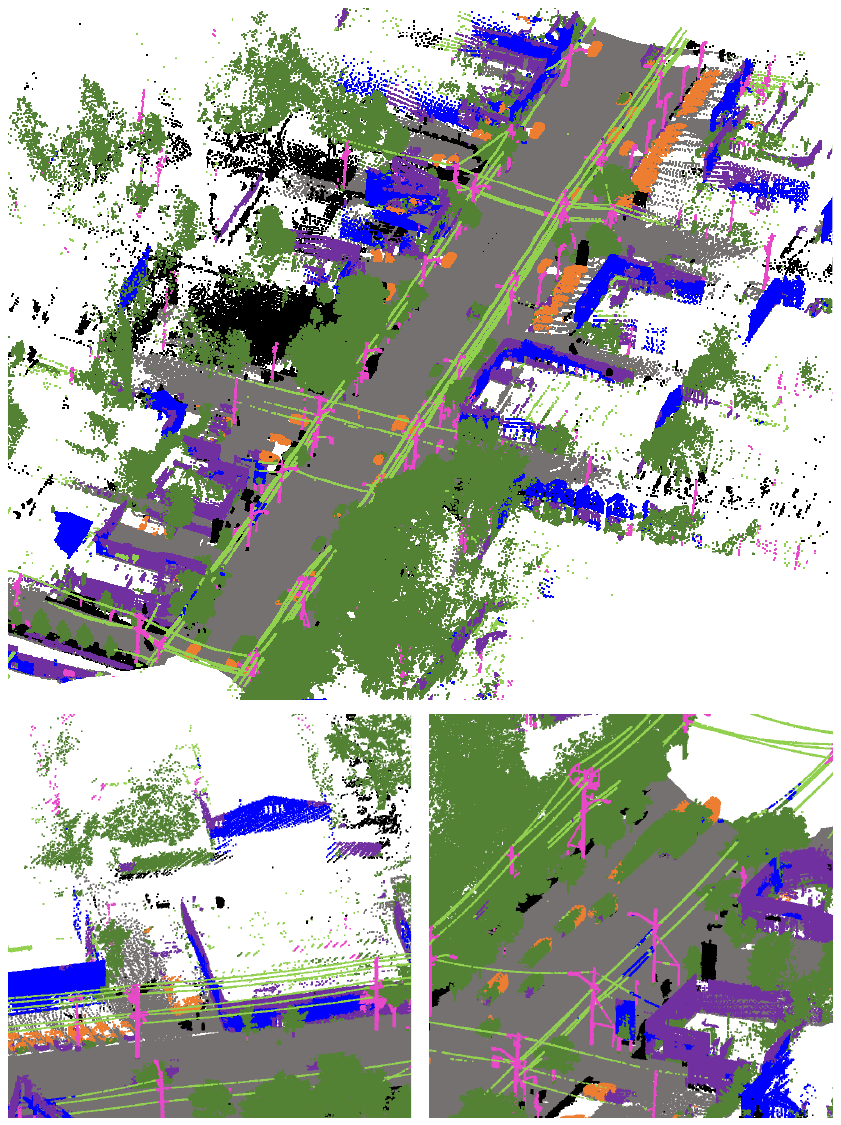}
    \caption{Source only}
\end{subfigure}
\hfill
\begin{subfigure}{0.32\linewidth}
    \centering
    \includegraphics[width=1\linewidth]{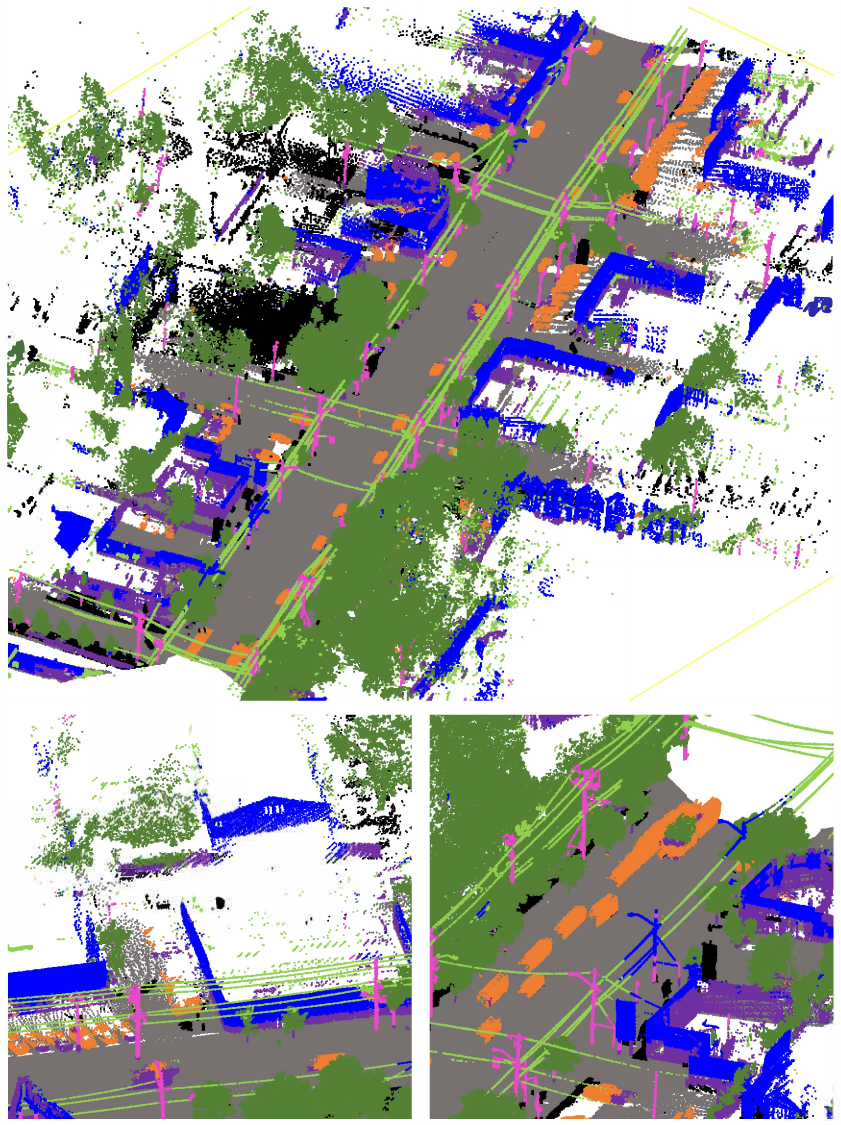}
    \caption{Our method}
\end{subfigure}
\hfill
\begin{subfigure}{0.32\linewidth}
    \centering
    \includegraphics[width=1\linewidth]{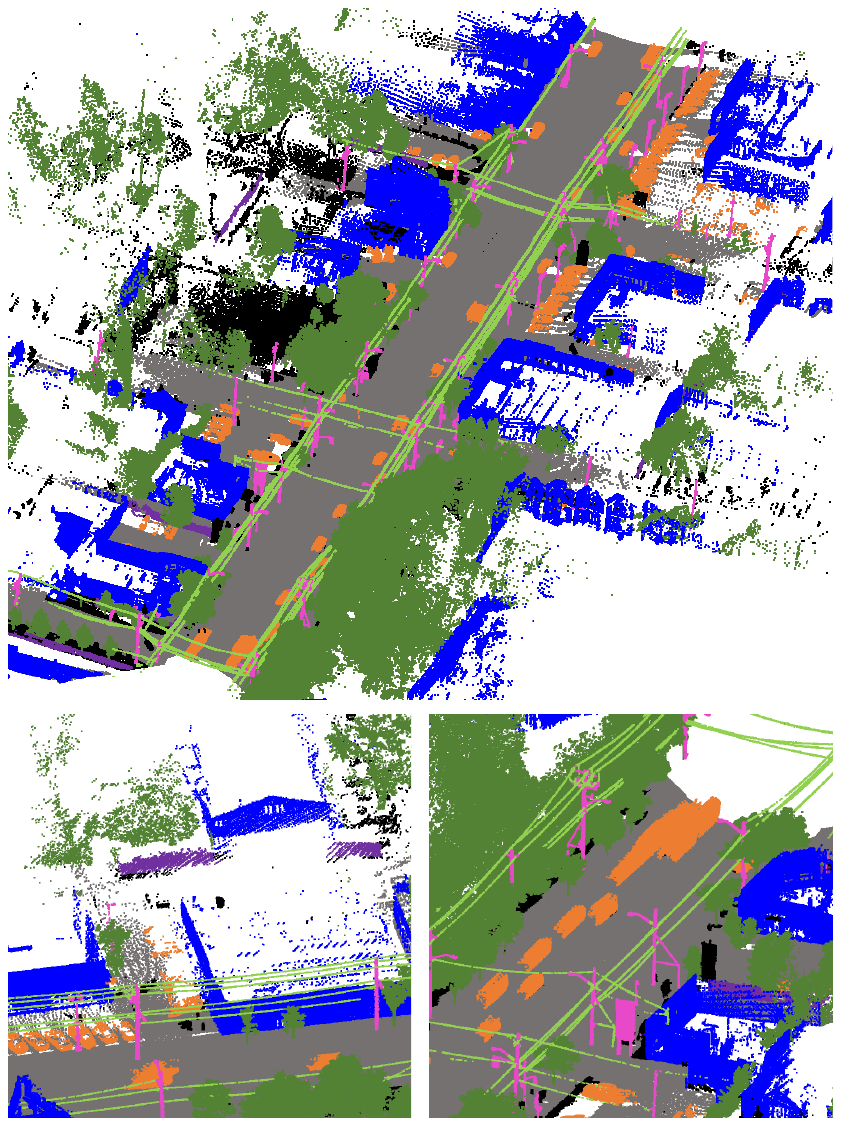}
    \caption{Ground truth}
\end{subfigure}
\caption{Classification result of T3D dataset with a model adapted from DALES dataset.}
\label{fig:res_dales_t3d}
\end{figure*}

\begin{table*}[b!]
\caption{Comparison of TTA methods on DALES to T3D dataset (\%)}
\label{tab:dales_t3d}
\centering
\small
\begin{tabularx}{\textwidth}{XXXXXXXXXX}
\hline
\multirow{2}{*}{Method} & \multicolumn{7}{c}{IoU} & \multirow{2}{*}{mIoU} & \multirow{2}{*}{OA} \\ 
\cline{2-8}  & Road & Natural & Building & Utility line & Pole & Car & Fence  \\ \hline                           
Source & 97.87 & 80.72 & 36.13 & 71.40 & 65.19 & 31.76 & 3.18 & 55.18 & 90.75 \\
AdaBN & 97.80 & 67.30 & 21.31 & 66.89 & 49.54 & 31.13 & 15.72 & 49.96 & 89.24 \\ 
TENT & 97.82 & 66.60 & 18.79 & 67.32 & 49.05 & 30.71 & 16.12 & 49.49 & 88.99  \\
DIGA & 98.01 & 88.49 & 50.71 & 73.33 & 67.86 & 66.32 & 3.95 & 64.10 & 93.18 \\
Ours  & 98.09 & 86.99 & 53.50 & 60.12 & 60.34 & 72.72 & 3.07 & 62.12 & 92.67 \\ 
\hline
\end{tabularx}
\end{table*}

\subsection{SensatUrban to H3D adaptation}

The classification map can be seen in Fig.~\ref{fig:res_h3d}. In the source mode, many vegetation points are incorrectly classified as urban furniture. Unlike the detailed tree canopy information captured by LiDAR sensors, photogrammetric point clouds capture only the surface, resulting in notable differences. Moreover, much of the data near the marina is misclassified as buildings. When our method is adapted, most tree points are correctly retrieved, and the misclassification of building points is reduced. The results of various methods are further quantified (see Table~\ref{tab:h3d}). Compared to the source mode, all adaptation methods significantly improve vegetation classification accuracy. Although other methods continue to perform poorly in building extraction, our method improves IoU by 18\%. We achieve a score of 59.46\% on mIoU and 85.97\% on OA, which significantly surpasses other adaptation approaches. It is worth noting that, with the inclusion of class prototypes, DIGA performs worse than AdaBN. DIGA reassigns labels by computing the distance between features and the average embedding of each class, which can be detrimental when inter-class boundary is not well defined.

\begin{figure*}[t!]
\centering
\begin{subfigure}[b]{1.0\linewidth}
    \centering
    \includegraphics[width=0.45\linewidth]{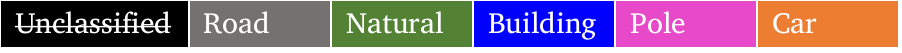}
\end{subfigure}
\hfill
\begin{subfigure}{0.32\linewidth}
    \centering
    \includegraphics[width=1\linewidth]{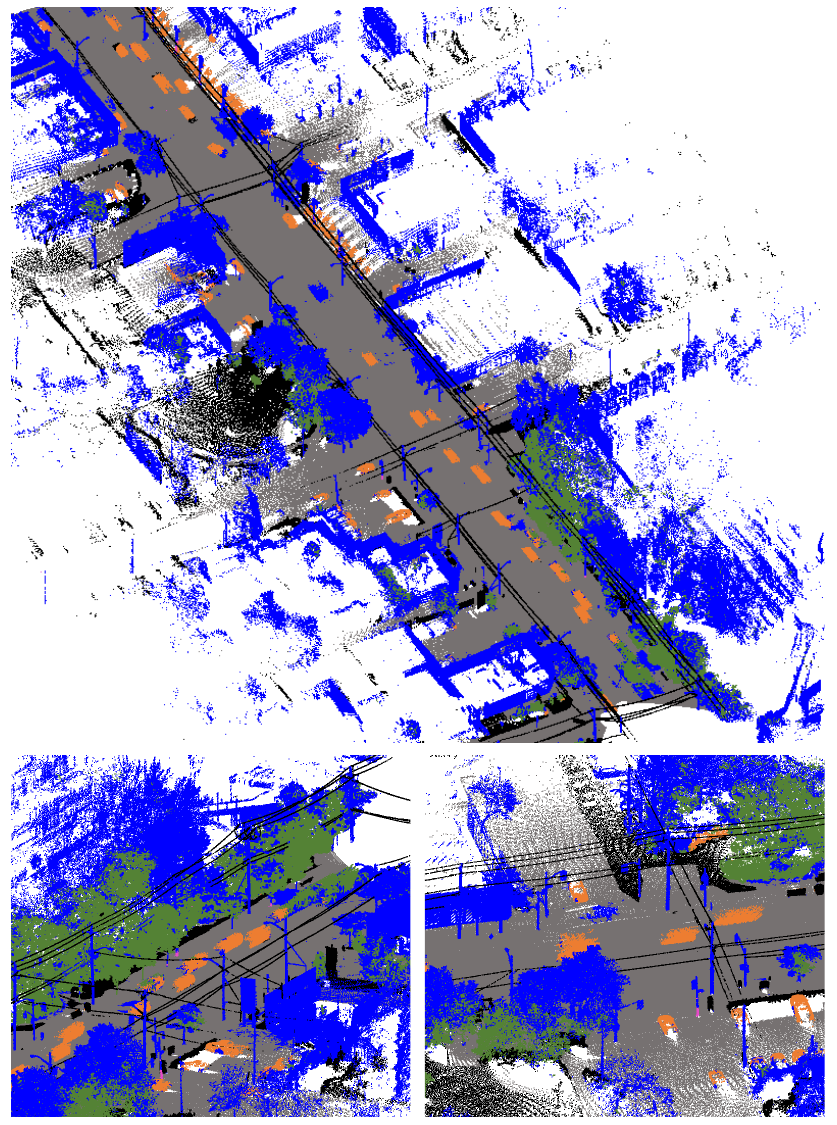}
    \caption{Source only}
\end{subfigure}
\hfill
\begin{subfigure}{0.32\linewidth}
    \centering
    \includegraphics[width=1\linewidth]{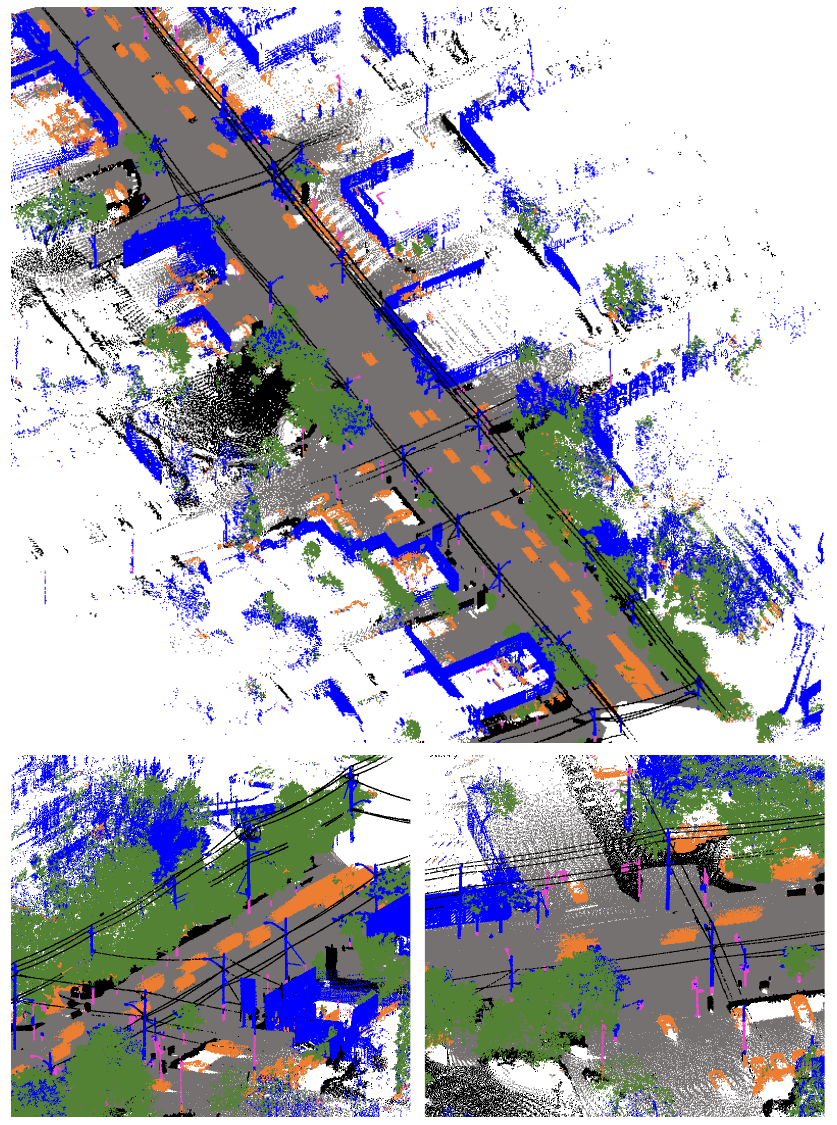}
    \caption{Our method}
\end{subfigure}
\hfill
\begin{subfigure}{0.32\linewidth}
    \centering
    \includegraphics[width=1\linewidth]{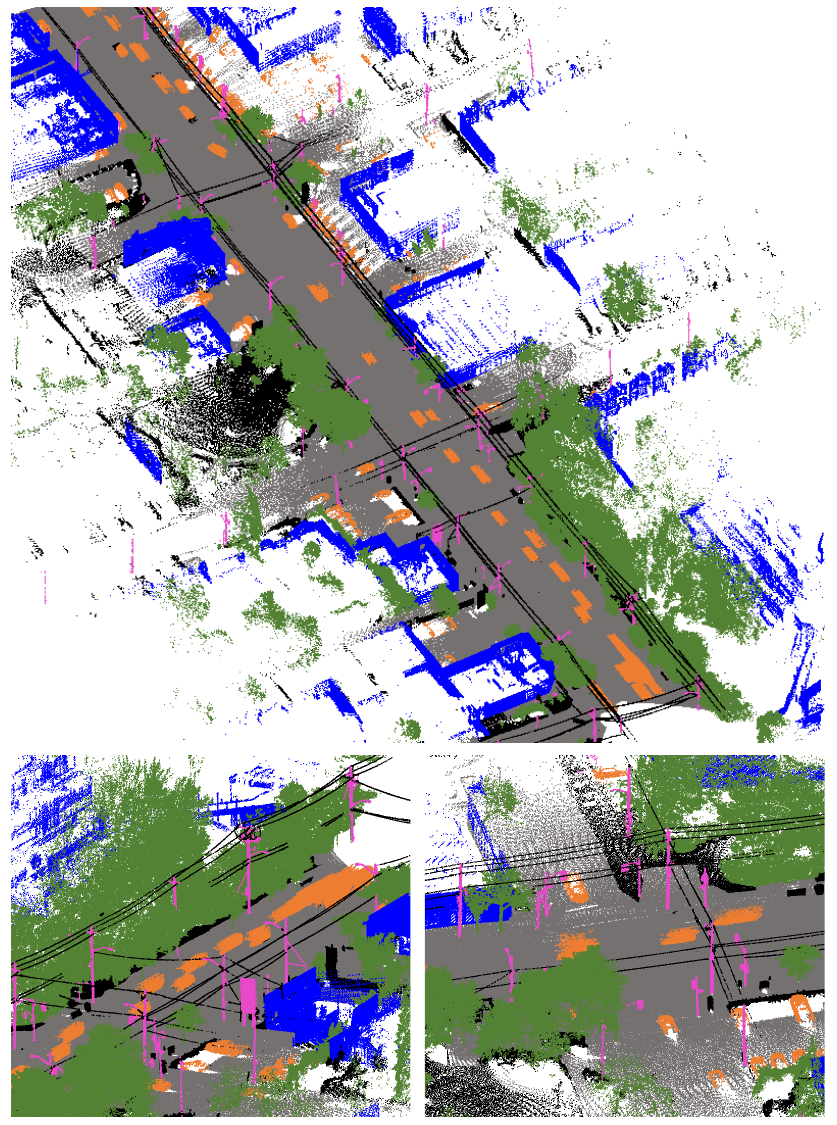}
    \caption{Ground truth}
\end{subfigure}
\caption{Classification result of T3D dataset with a model adapted from SynthCity dataset.}
\label{fig:res_syn_t3d}
\end{figure*}

\begin{table*}[t!]
\caption{Comparison of TTA methods on SynthCity to T3D dataset (\%)}
\label{tab:syn_t3d}
\centering
\small
\begin{tabularx}{\textwidth}{XXXXXXXX}
\hline
\multirow{2}{*}{Method} & \multicolumn{5}{c}{IoU} & \multirow{2}{*}{mIoU} & \multirow{2}{*}{OA} \\ 
\cline{2-6}  & Road & Natural & Building & Pole & Car \\ \hline                           
Source & 96.97 & 54.47 & 41.10 & 1.45 & 39.58 & 46.71 & 86.26 \\
AdaBN & 93.67 & 59.58 & 38.22 & 10.76 & 37.38 & 47.92 & 84.96 \\ 
TENT & 93.39 & 60.58 & 38.05 & 10.51 & 36.69 & 47.84 & 84.91 \\ 
DIGA & 98.73 & 69.45 & 51.57 & 11.95 & 41.59 & 54.66 & 90.66 \\
Ours & 98.79 & 84.03 & 71.32 & 19.69 & 63.11 & 67.39 & 94.73 \\ 
\hline
\end{tabularx}
\end{table*}

\subsection{DALES to T3D adaptation}
We illustrate the classification result in Fig.~\ref{fig:res_dales_t3d}. The source mode effectively classifies the points on the road, pole, and utility line because these objects are less affected by the scanning viewpoints. For the building class, the absent roof in the MLS data causes a severe confusion when adapting from ALS data. Furthermore, car structure scanned from the overhead is also not as rich as that from street view perspective, leading to incorrect predictions. Our method effectively recalls incorrectly classified building and car points. The quantitative comparison is shown in Table~\ref{tab:dales_t3d}. Our method achieves an increase 7\% in mIoU, reaching 62.12\%. In contrast, AdaBN and TENT even obtain an inferior result in comparison to the source mode. This means that directly replacing BN statistics with values calculated from each testing batch can produce unstable predictions. In this experiment, DIGA obtains the highest mIoU number, which also slightly outperforms our method in terms of OA.

\subsection{SynthCity to T3D adaptation}
The classification results are shown in Fig.~\ref{fig:res_syn_t3d}. Due to the substantial differences between synthetic and real-world data concerning geometric structure and color rendering, the source model incorrectly classifies most points as buildings, except for ground and vehicle points. Through the enhancement of our technique, the pre-trained model adapts to the limited positive feedback, resulting in a more balanced classification pattern. Our method not only effectively retains building points but also successfully reclassifies a significant number of previously misclassified natural and pole points. According to the quantitative evaluation comparison (see Table~\ref{tab:syn_t3d}), our method shows a substantial performance improvement over the source model, with a 21\% increase in mIoU. Conversely, AdaBN and TENT face challenges in adapting the pre-trained model, leading to performance declines. DIGA significantly increases the classification accuracy compared to the source model, yet it remains inferior to our results. This indicates that a meticulously designed self-supervision strategy outperforms back-free methods when adapting pre-trained models for geospatial PCSS in cases of large domain shifts.

\section{Discussion}
\label{sec:dis}

\subsection{Ablation study}

The performance of each module in the proposed method is evaluated here, and the quantitative results are shown in Table~\ref{tab:ablation}. PBN significantly boosts OA and mIoU in transitioning from SensatUrban to H3D (S2H) and from SynthCity to T3D (S2T) when compared to the baseline. In the case of DALES to T3D (D2T) adaptation, even though AdaBN has worse results, our proposed PBN still retains the original effectiveness. Furthermore, integrating IM shows a substantial improvement in evaluation metrics for both S2H and S2T. Specifically, for S2T adaptation, there is an approximate 20\% increase in mIoU. It is important to mention that there is notable performance decline in D2T, indicating the necessity of PL. Further performance enhancements are observed when combining PL, with results suggesting that reliable auxiliary supervisory signals offer an effective path for model adaptation. These experimental results demonstrate that each module individually improves classification performance and that the combined version achieves the best results.

\begin{table*}[t!]
\caption{Ablation study (\%)}
\label{tab:ablation}
\centering
\small
\setlength\tabcolsep{15pt}%
\begin{tabularx}{\linewidth}{cccXXXXXX}
\hline
\multicolumn{3}{c}{Module} & \multicolumn{2}{l}{SensatUrban $\Rightarrow$ H3D} & \multicolumn{2}{l}{DALES $\Rightarrow$ T3D} & \multicolumn{2}{l}{SynthCity $\Rightarrow$ T3D} \\ \hline
PBN & IM & PL & mIoU & OA & mIoU & OA & mIoU & OA \\ \hline
&     &      & 41.94 & 73.53 & 55.18 & 90.75 & 46.71 & 86.26 \\
\checkmark & & & 48.78 & 77.67 & 54.98 & 90.85 & 50.17 & 88.61 \\
\checkmark & \checkmark & & 54.95 & 82.28 & 48.32 & 87.98 & 65.91 & 93.43 \\
\checkmark & \checkmark & \checkmark & 59.46 & 85.97 & 62.12 & 92.67 & 67.39 & 94.73 \\
\hline
\end{tabularx}
\end{table*}

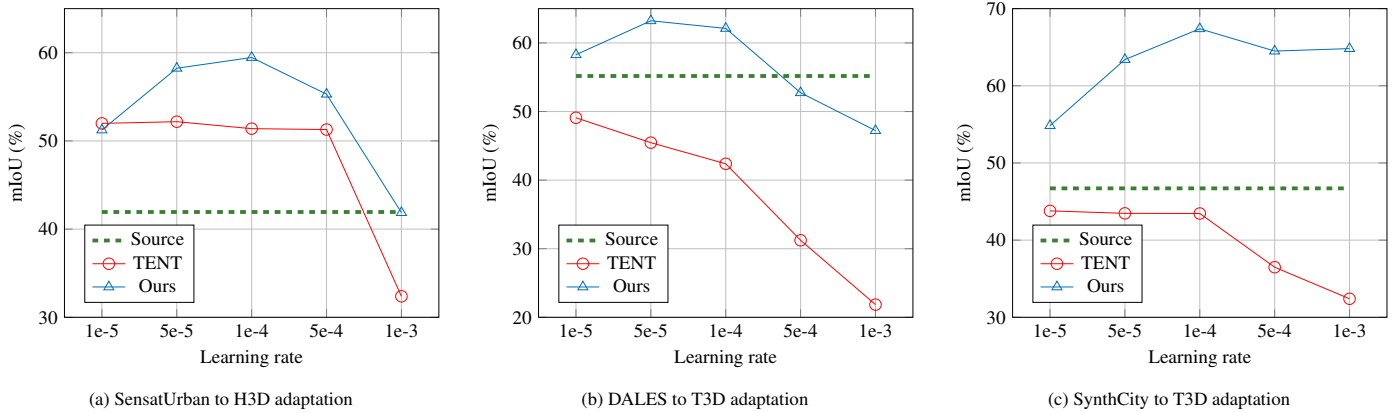
\begin{figure*}[t]
\centering
    \begin{subfigure}[b]{0.32\textwidth}
    \centering
    
    \resizebox{1.0\textwidth}{!}{
    \begin{tikzpicture}
    \begin{axis}[
    grid=both,
    ymin=30, ymax=65,
    xmin=0.5, xmax=5.5,
    xtick={1, 2, 3, 4, 5},
    xticklabels={1e-5, 5e-5, 1e-4, 5e-4, 1e-3},
    xlabel=Learning rate, 
    ylabel=mIoU (\%), 
    legend style={at={(0.2,0.3)},anchor=north},
    ]

    \addplot[dashed, OliveGreen, line width=2pt, domain=1:5] {41.94};
    \addlegendentry{Source}
    
    \addplot[mark=o, mark size=3pt, red] plot coordinates 
    { 
     (1,51.99)
     (2,52.18)
     (3,51.39)
     (4,51.29)
     (5,32.38)
    };
    \addlegendentry{TENT}
 
    \addplot[mark=triangle, mark size=3pt, NavyBlue] plot coordinates {
     (1,51.24)
     (2,58.24)
     (3,59.46)
     (4,55.30)
     (5,41.87)
    };
    \addlegendentry{Ours}
    \end{axis}
    \end{tikzpicture}
    }
    \caption{SensatUrban to H3D adaptation}
    \end{subfigure}
    \hfill
     \begin{subfigure}[b]{0.32\textwidth}
    \centering
    
    \resizebox{1.0\textwidth}{!}{
    \begin{tikzpicture}
    \begin{axis}[
    grid=both,
    ymin=20, ymax=65,
    xmin=0.5, xmax=5.5,
    xtick={1, 2, 3, 4, 5},
    xticklabels={1e-5, 5e-5, 1e-4, 5e-4, 1e-3},
    xlabel=Learning rate, 
    ylabel=mIoU (\%), 
    legend style={at={(0.2,0.3)},anchor=north},
    ]

    \addplot[dashed, OliveGreen, line width=2pt, domain=1:5] {55.18};
    \addlegendentry{Source}
    
    \addplot[mark=o, mark size=3pt, red] plot coordinates 
    { 
     (1,49.09)
     (2,45.46)
     (3,42.38)
     (4,31.23)
     (5,21.85)
    };
    \addlegendentry{TENT}
 
    \addplot[mark=triangle, mark size=3pt, NavyBlue] plot coordinates {
     (1,58.28)
     (2,63.23)
     (3,62.12)
     (4,52.74)
     (5,47.22)
    };
    \addlegendentry{Ours}
    \end{axis}
    \end{tikzpicture}
    }
    \caption{DALES to T3D adaptation}
    \end{subfigure}
    \hfill
    \begin{subfigure}[b]{0.32\textwidth}
    \centering
    
    \resizebox{1.0\textwidth}{!}{
    \begin{tikzpicture}
    \begin{axis}[
    grid=both,
    ymin=30, ymax=70,
    xmin=0.5, xmax=5.5,
    xtick={1, 2, 3, 4, 5},
    xticklabels={1e-5, 5e-5, 1e-4, 5e-4, 1e-3},
    xlabel=Learning rate, 
    ylabel=mIoU (\%), 
    legend style={at={(0.2,0.3)},anchor=north},
    ]

    \addplot[dashed, OliveGreen, line width=2pt, domain=1:5] {46.71};
    \addlegendentry{Source}
    
    \addplot[mark=o, mark size=3pt, red] plot coordinates 
    { 
     (1,43.79)
     (2,43.47)
     (3,43.45)
     (4,36.49)
     (5,32.40)
    };
    \addlegendentry{TENT}
 
    \addplot[mark=triangle, mark size=3pt, NavyBlue] plot coordinates {
     (1,54.84)
     (2,63.40)
     (3,67.39)
     (4,64.50)
     (5,64.82)
    };
    \addlegendentry{Ours}
    \end{axis}
    \end{tikzpicture}
    }
    \caption{SynthCity to T3D adaptation}
    \end{subfigure}

\caption{Robustness analysis on learning rate}
\label{fig:lr}
\end{figure*}

\subsection{Influence of learning rate}
The learning rate plays a crucial role when self-supervised techniques are incorporated into TTA using unlabeled target domain data. Unlike traditional supervised deep model training, a smaller learning rate is required to ensure effective adaptation and prevent model collapse. However, a learning rate that is too small may not lead to significant performance improvements. Therefore, it is important to assess the robustness of methods across different learning rate values. The comparison between our approach and TENT is shown in Fig.~\ref{fig:lr}. The figure demonstrates that choosing an optimal learning rate enables successful adaptation, enhancing performance effectively while avoiding negative impacts. A very small learning rate results in only minor model updates, insufficient for meaningful performance improvements. However, an excessively high learning rate can produce large gradients that misdirect model updates, causing model collapse. Our experiments indicate that a learning rate of 1e-4 yields the best results in most cases. For TENT, only a very low learning rate yields average performance with the ER constraint, whereas a high learning rate significantly reduces accuracy.

\subsection{Robustness analysis}
During inference, we leverage overlapping sub-clouds, which has been shown to be effective in established point cloud backbone networks. With model parameters being fine-tuned through sub-cloud generation, we analyze the effect of random mini-batch creation and testing sequence. We record the mean performance and the corresponding standard deviation across five inference runs. As shown in Table~\ref{tab:robust}, our method produces consistent and robust results, evidencing its resilience to mini-batch and testing sequence variations. Across the three adaptation experiments, our method demonstrates a low standard deviation for both mIoU and OA, underscoring its robustness.

\begin{table}[h]
\caption{Robustness analysis of the proposed method}
\label{tab:robust}
\centering
\small
\setlength\tabcolsep{3pt}%
\begin{tabularx}{\linewidth}{cXXXXXX}
\hline
\multirow{2}{*}{} & \multicolumn{2}{c}{SensatUrban $\Rightarrow$ H3D} & \multicolumn{2}{c}{DALES $\Rightarrow$ T3D} & \multicolumn{2}{c}{SynthCity $\Rightarrow$ T3D} \\ 
\cline{2-7} 
& mIoU(\%) & OA(\%) & mIoU(\%) & OA(\%) & mIoU(\%) & OA(\%)
\\ \hline
Iter1  & 59.46  & 85.97  & 62.12  & 92.67  & 67.39  & 94.73  \\
Iter2  & 58.73  & 85.62  & 62.90  & 93.19  & 65.92  & 93.95  \\
Iter3  & 60.56  & 87.32  & 61.75  & 92.72  & 66.95  & 94.71  \\
Iter4  & 60.97  & 87.32  & 64.77  & 94.15  & 66.20  & 94.20  \\
Iter5  & 59.69  & 86.05  & 62.36  & 93.44  & 65.74  & 94.37  \\
\hline
\textbf{Mean}  & 59.88  & 86.46  & 62.78  & 93.23  & 66.44  & 94.37  \\
\textbf{STD}  & 0.89  & 0.81  & 1.19  & 0.37  & 0.70  & 0.34  \\  \hline
\end{tabularx}
\end{table}

\subsection{Influence of model parameter adaptation schemes}

In this research, we choose to update the BN layers in deep learning models to ensure stable adaptation. To evaluate the effectiveness of our method, we conducted a comparative study with two other model parameter adaptation strategies. The first strategy, called the `All' mode, involves updating all learnable parameters in the model. The second strategy, which is based on SHOT \citep{10.5555/3524938.3525498}, separates the classifier (the final fully connected layer) from the feature extractor (FE, consisting of all previous layers). According to SHOT, the FE mainly encodes distributional data and thus targets the updating of FE parameters. The results of our performance comparison are presented in Table~\ref{tab:param}. The analysis shows that both the `All' and `FE' modes are significantly less effective than focusing solely on the BN layers. Specifically, in D2T adaptation, using the `All' and `FE' modes results in model collapse. This comparison reinforces the idea that BN layers hold domain-specific information, which also applies to point cloud processing.

\begin{table}[h]
\caption{Performance comparison under different model parameter adaptation schemes (\%)}
\label{tab:param}
\centering
\small
\setlength\tabcolsep{3pt}%
\begin{tabularx}{\linewidth}{cXXXXXX}
\hline
\multirow{2}{*}{} & \multicolumn{2}{c}{SensatUrban $\Rightarrow$ H3D} & \multicolumn{2}{c}{DALES $\Rightarrow$ T3D} & \multicolumn{2}{c}{SynthCity $\Rightarrow$ T3D} \\ 
\cline{2-7} 
& mIoU & OA & mIoU & OA & mIoU & OA
\\ \hline
Source  & 41.94 & 73.53 & 55.18 & 90.75 & 46.71 & 86.26 \\
All  & 47.66 & 80.74 & 21.72 & 66.23 & 59.14 & 89.62 \\
FE  & 46.07 & 79.73 & 24.11 & 69.04 & 56.24 & 89.70 \\
BN  & 59.46  & 85.97  & 62.12  & 92.67  & 67.39  & 94.73 \\ \hline
\end{tabularx}
\end{table}

\begin{table}[h]
\caption{Comparison of TTA and full supervision scheme (\%)}
\label{tab:full}
\centering
\small
\setlength\tabcolsep{3pt}%
\begin{tabularx}{\linewidth}{cXXXXXX}
\hline
\multirow{2}{*}{} & \multicolumn{2}{c}{SensatUrban $\Rightarrow$ H3D} & \multicolumn{2}{c}{DALES $\Rightarrow$ T3D} & \multicolumn{2}{c}{SynthCity $\Rightarrow$ T3D} \\ 
\cline{2-7} 
& mIoU & OA & mIoU & OA & mIoU & OA
\\ \hline
Source  & 41.94 & 73.53 & 55.18 & 90.75 & 46.71 & 86.26 \\
Ours  & 59.46  & 85.97  & 62.12  & 92.67  & 67.39  & 94.73 \\ \hline
Full  & 73.94  & 93.78  & 85.13  & 98.65  & 93.61  & 98.84  \\  \hline
\end{tabularx}
\end{table}

\begin{table*}[t!]
\caption{Performance comparison under different backbone networks (\%)}
\label{tab:backbone}
\centering
\small
\begin{tabularx}{\linewidth}{XXXXXXXX}
\hline
\multirow{2}{*}{} & & \multicolumn{2}{l}{SensatUrban $\Rightarrow$ H3D} & \multicolumn{2}{l}{DALES $\Rightarrow$ T3D} & \multicolumn{2}{l}{SynthCity $\Rightarrow$ T3D} \\ 
\cline{3-8} 
& & mIoU & OA & mIoU & OA & mIoU & OA
\\ \hline
\multirow{2}{*}{KPConv} & Source  & 41.94 & 73.53 & 55.18 & 90.75 & 46.71 & 86.26 \\
& Ours  & 59.46  & 85.97  & 62.12  & 92.67  & 67.39  & 94.73 \\ \hline
\multirow{2}{*}{PointNet++} & Source  & 52.94  & 82.89  & 42.28  & 86.93  & 47.76  & 86.81  \\  
& Ours  & 55.71  & 84.96  & 50.57  & 87.47  & 54.21  & 89.25 \\ \hline
\end{tabularx}
\end{table*}

\subsection{Comparison with the full supervision scheme}

Without the availability of labeled target data, TTA focuses on streamlining the computation process and enhancing time efficiency, though its performance isn't anticipated to match that under full supervision. Here, we examine the performance disparity between our approach and the fully supervised framework, which adheres to the standard training and testing procedures on the original benchmark. It should be emphasized that optimal results can be achieved with full supervision. For instance, in the D2T scenario, only 3D coordinate data is used for inference since the DALES dataset lacks color information, while the T3D dataset includes it. The quantitative results are shown in Table~\ref{tab:full}. This study's benchmark consolidates similar categories due to cross-domain class discrepancies, resulting in very high evaluation metrics across all three experiments in the fully supervised setting. Although our method significantly enhances accuracy over the source model, the figures remain notably lower than those under full supervision. Nevertheless, TTA's practicality reveals potential in addressing domain-shift challenges. To further boost performance, integrating weakly labeled target data could be a promising area for future research.

\subsection{Influence of backbone network}

Our method introduces a flexible framework that can be integrated with various deep networks. To evaluate its flexibility, we employ PointNet++ \citep{NIPS2017_d8bf84be} as an additional backbone for comparison. As shown in Table~\ref{tab:backbone}, our method considerably enhances accuracy compared to the source model with two distinct backbones. For D2T adaptation, despite PointNet++ initially displaying subpar results, our method rectifies erroneous predictions and significantly improves evaluation metrics. KPConv and PointNet++ exhibit marked performance differences across three benchmarks, indicating that the architecture of deep models may also be crucial to adaptation performance. This suggests an interesting direction for future investigation.

\section{Conclusion}
\label{sec:con}

In this study, we investigate test-time adaptation for point cloud semantic segmentation, which is required to directly adapt a pre-trained model to a target domain with distinct characteristic discrepancy during inference time. To achieve this objective, we designed a TTA framework targeted at BN layers, since they are regarded to contain fruitful domain-specific features. A progressive batch normalization (PBN) adaptation module is proposed by adapting statistical information of BN layers with that calculated from testing batches. Additionally, to achieve complete BN adaptation, we introduce a self-supervised module containing information maximization and pseudo-labeling to optimize learnable parameters in BN layers. To evaluate the effectiveness of our method, we develop three practical adaptation modes, including photogrammetric to ALS data, ALS to MLS data, and synthetic to MLS data. Our method significantly improves OA and mIoU compared to the source-only mode, which also outperforms the existing popular TTA counterparts. Evaluated on the SensatUrban to Hessigheim 3D dataset, our method achieved an overall accuracy of 85.97\% and a mIoU of 59.46\%, which increased by approximately 12.5\% and 17.5\%, respectively, compared to baseline. In the future, we will explore the potential of large vision-language models in geospatial PCSS and achieve open-set adaptation.

\section*{Acknowledgements}

This work was supported by National Natural Science Foundation of China (Grant No.42171361) and by a grant from the Research Grants Council of the Hong Kong Special Administrative Region, China (Project No. PolyU 15215023). 




\bibliographystyle{elsarticle-harv} 
\bibliography{ref}





\end{document}